\documentclass[submit]{elsarticle}

\usepackage{lineno,hyperref}
\usepackage{amsmath,amssymb,amsfonts}
\usepackage{algorithmic}
\usepackage[algo2e]{algorithm2e} 
\usepackage{algorithm}
\usepackage{graphicx}
\usepackage{textcomp}
\usepackage{xcolor}
\usepackage{multirow}
\usepackage{tabularx}
\newcolumntype{Y}{>{\centering\arraybackslash}c}

\usepackage{amsthm}
  {
      \theoremstyle{plain}

      \newtheorem{remark}{Remark}
      \newtheorem{proposition}{Proposition}

  }

\usepackage{amsmath}
\usepackage{soul}
\usepackage{subfigure}

\usepackage{caption}
\modulolinenumbers[5]

\journal{arXiv}

\renewcommand{\linenumbers}{}
\begin{document}
\nolinenumbers
\begin{frontmatter}

\title{Technical Report: Assisting Backdoor Federated Learning with Whole Population Knowledge Alignment}

\author{Tian Liu}
\ead{tianliu@auburn.edu}
\author{Xueyang Hu}
\ead{xzh0051@auburn.edu}
\author{Tao Shu \corref{cor1}}
\ead{tshu@auburn.edu}
\address{ Department of Computer Science and Software Engineering, Auburn University, Auburn, AL, USA}
\cortext[cor1]{Corresponding author}




\begin{abstract}
Due to the distributed nature of Federated Learning (FL), researchers have uncovered that FL is vulnerable to backdoor attacks, which aim at injecting a sub-task into the FL without corrupting the performance of the main task. Single-shot backdoor attack achieves high accuracy on both the main task and backdoor sub-task when injected at the FL model convergence. However, the early-injected single-shot backdoor attack is ineffective because: (1) the maximum backdoor effectiveness is not reached at injection because of the dilution effect from normal local updates; (2) the backdoor effect decreases quickly as the backdoor will be overwritten by the newcoming normal local updates. In this paper, we strengthen the early-injected single-shot backdoor attack utilizing FL model information leakage. We show that the FL convergence can be expedited if the client trains on a dataset that mimics the distribution and gradients of the whole population. Based on this observation, we proposed a two-phase backdoor attack, which includes a preliminary phase for the subsequent backdoor attack. In the preliminary phase, the attacker-controlled client first launches a whole population distribution inference attack and then trains on a locally crafted dataset that is aligned with both the gradient and inferred distribution. Benefiting from the preliminary phase, the later injected backdoor achieves better effectiveness as the backdoor effect will be less likely to be diluted by the normal model updates. Extensive experiments are conducted on MNIST dataset under various data heterogeneity settings to evaluate the effectiveness of the proposed backdoor attack. Results show that the proposed backdoor outperforms existing backdoor attacks in both success rate and longevity, even when defense mechanisms are in place.
\end{abstract}


\end{frontmatter}

\linenumbers

\section{Introduction}
Federated learning (FL) \cite{konevcny2016federated, mcmahan2017communication} is a distributed learning system, which allows multiple clients to collaboratively train a high accuracy model by taking advantage of a wide range of data from physically separated clients without sharing their locally collected data. Currently, FL applications thrive in next-word and emoji prediction on smartphones \cite{chen2019federated, yang2018applied, ramaswamy2019federated, hard2018federated}, environmental monitoring \cite{han2019visual}, and aiding in medical diagnosis among hospitals \cite{xu2020federated, brisimi2018federated}. 

Due to the distributed nature and inherent data heterogeneity (i.e., data being non-i.i.d.) across FL clients, the local model updates uploaded by clients may be different from others. As a result, it is assumed that the central server is unable to validate the legitimacy of the received model updates, which provides a venue for new attacks. Backdoor is one of the data poisoning attacks \cite{bagdasaryan2018backdoor}, in which an adversary corrupts the global model such that the new global model reaches a high accuracy on the FL main task, as well as on a backdoor sub-task activated by some trigger, and such a high backdoor sub-task accuracy retains for multiple training rounds. The backdoor attack has been shown unavoidable and computationally hard to detect \citep{wang2020attack}.

The single-shot backdoor, in which the adversary has only one chance to inject the designed backdoor trigger, achieves a high success rate and long-lasting longevity when being injected as the FL model is close to convergence. However, in practice, a malicious client cannot manipulate the timing of the backdoor injection because clients are randomly selected to participate in a round of training. It is not guaranteed that the backdoor is always injected at the time of convergence. Compared to those injected at the convergence of the FL model, the single-shot backdoor injected in the early training stage (before the FL model converges) is weaker due to the following two reasons. Firstly, the backdoor effect is not maximized at injection because of the dilution effect of local updates from normal clients. The backdoor attack accomplishes its goal by replacing the global model with a scaled backdoored local one, such that the corrupted FL global model achieves high accuracy on both the main task and backdoor sub-task. However, the effectiveness of the replacement is under the dilution of model updates from normal clients. Therefore, the maximum effectiveness of the backdoor is usually reached at the convergence of the FL model, where the magnitude of (normal) local updates goes to 0 and local updates cancel out in the server aggregation. For single-shot backdoors injected in a very early stage, the large magnitude of model updates from other clients generates a great dilution effect, and such an effect will undermine the strength of the backdoored local model update. Secondly, the backdoor effect decreases quickly as the injected backdoor will be overwritten by newcoming model updates from normal clients. In general, the magnitude of local model updates decreases along the FL training course. As a result, the earlier the backdoor is injected, the faster the backdoor effect will diminish. Despite the ineffectiveness of early-injected backdoors, the early training stage is still vulnerable to backdoor attacks, as participating clients may quit early when the expected model accuracy is reached.

Meanwhile, another branch of research demonstrates that although clients' private data is not directly revealed, the shared FL global model can unintentionally leak sensitive information about the data on which it was trained \cite{dwork2010difficulties}. As pointed out by previous studies, the attacker can reconstruct the training data \cite{fredrikson2015model}, deanonymize participants \cite{orekondy2018gradient}, and even infer class representatives \cite{hitaj2017deep} and data sample membership \cite{melis2019exploiting, nasr2018comprehensive}.

The ineffectiveness of the early-injected single-shot backdoor attack combined with such privacy issues in FL motivates us to consider the following research problem: \textbf{does FL information leakage render a stronger backdoor attack?} Our main insight is that the weight divergence \cite{xu2020federated}, which results in slow and unstable convergence of the FL model, is introduced by the difference of the label distribution (henceforth referred to as ``distribution") and the difference of gradients between a single client's local data and the whole population, namely the aggregation of all clients' data. Reducing such distances can expedite FL convergence and overcome the aforementioned deficiency of early-injected single-shot backdoors.

\textbf{In this paper, we propose a novel information leakage assisted two-phase FL backdoor attack, which enhances the effectiveness of FL early-injected single-shot backdoor attack.} We assume the attacker-controlled clients can interact with the FL training for multiple times, but they have only one chance to inject the backdoor. In our design, we do not directly strengthen the backdoor attack. Instead, we design a preliminary phase for the subsequent backdoor injection, where the attacker-controlled clients play the role of accomplices and reach out the FL global model by uploading model updates that are beneficial to FL convergence, so as to pave the way for the subsequent backdoor injection. Formally, the proposed backdoor attack consists of two phases: a preliminary phase, in which the attacker-controlled clients help to accelerate the convergence of the FL model, and an attack phase, in which the backdoor attack is launched. In the preliminary phase, the attacker-controlled clients first perform a passive inference attack to get an estimate of the whole population distribution. Then, instead of training on the original local data, they train on locally crafted datasets, whose distributions are aligned with the inferred whole population distribution, so that the weight divergence is reduced and the FL model converges more quickly. Although the operations in the preliminary phase seem legitimate, they help to improve the effectiveness and persistence of the backdoor by reducing the dilution effect from other clients (as the magnitude of their local model updates decreases more quickly). 

When the expected FL model accuracy is reached or the client that has the capability to perform a backdoor attack is selected, the backdoor attacked is launched by training on a locally poisoned dataset and the backdoored local model updates are scaled up before submitting to the FL server. Benefiting from the preliminary phase, the single-shot backdoor injected in the resulting FL model will be less likely to be diluted by model updates from other clients. Therefore, the designed preliminary phase successfully overcomes the deficiencies of early-injected single-shot backdoor and significantly improves the strength and persistence of the backdoor effect. Note that the proposed preliminary phase benefits the backdoor effectiveness by improving the FL convergence, and hence reduces the dilution effect from other clients. And this preliminary phase is independent of the attack phase and therefore can be combined with any kind of backdoor attacks to enhance their backdoor performance.

To the best of our knowledge, we are the first in the literature to enhance the effectiveness of FL backdoor attacks by utilizing the information leaked from the FL model. Our \textbf{contributions} in this paper are four-folds:
\begin{itemize}
\item We prove an upper bound for the intra-aggregation weight divergence between the FL model and the centralized learning (CL) model, and demonstrate that the weight divergence is small. Thus, the FL global model updates can be used to approximate the CL model updates. 

\item We propose a novel optimization-based whole population distribution inference attack utilizing the above approximation and the linearity of the cross-entropy. Unlike the existing property inference attack, in which it can only generate binary property inference results, our proposed inference attack produces precise quantitative property information about the dataset. 

\item We propose a preliminary phase for the early-injected single-shot backdoor attack, which improves the attack effectiveness by reducing the dilution effect from local updates of normal clients. Specifically, attacker-controlled clients use the inferred distribution to craft auxiliary datasets using augmentation and downsampling techniques, such that the distribution of the auxiliary dataset is aligned with both the gradient and inferred global distribution. Training on the auxiliary dataset can facilitate the convergence of the FL model and reduce the magnitude of local model updates from normal clients, and further boost the performance of the backdoor attack.

\item Extensive experiments are conducted on MNIST dataset under various data heterogeneity settings to evaluate the accuracy of the proposed whole population distribution inference attack, the improvement of the convergence of the FL global model brought about by the proposed preliminary phase, as well as the effectiveness of the proposed backdoor attack. We also evaluate the proposed attack against two state-of-the-art defense mechanisms. The experimental results show that the proposed inference attack achieves high accuracy against FL in scenarios with and without defense mechanisms. The FL model assisted by the preliminary phase has a faster convergence rate, especially in the early training stage. The proposed backdoor outperforms existing backdoor attacks in both success rate and longevity, even when defense mechanisms are in place.
\end{itemize}

\textbf{Paper Organization.} This paper is structured as follows. We start by providing the background and related work in Section \ref{section::background and related work}. We present the threat model and attack design philosophy in Section \ref{section::Threat Model and Attack Design Philosophy}. Subsequently, the overview and detailed attack steps are presented in Section \ref{section::our_approach}. Finally, the experimental setup and results are presented in Sections \ref{section::experiments_setup} and \ref{section::results}, respectively. We evaluate the robustness of the proposed backdoor attack against two defense mechanisms in Section \ref{section::defense}, and we conclude our work in Section \ref{section::conclusion}.

Throughout this paper, we use the following notation:
\begin{itemize}
    \item $\| \cdot \|$ denotes the $\ell_{2}$ norm.
    \item $D_k$ and $D$ denote the training data on the $k$-th client and the entire population of training data, respectively. And we have $D = \cup_{k=1}^N D_k$.
    \item $n_k$ and $n$ denote the number of training samples in $D_k$ and $D$, respectively. And we have $n = \sum_{k=1}^{K} n_k$. 
    \item $w_k^{T}$ and $w^{T}$ denote the $k$-th local model weight and the global model weight at the $T$-th aggregation, respectively. 
    \item $F_k(w_k; D_k)$ and $F(w; D)$ denote the loss function on the $k$-th client and the loss function of a CL model, respectively.
    \item $\nabla L(w_k; D_k)$ and $\nabla L(w; D)$ denote the gradients of loss of the client $k$ and the gradients of loss of the CL model, respectively.
    \item $p(y = c)$ is the proportion of the label $c$ in the training data, and we have $\sum_{c=1}^{C} p(y=c) =1$.
\end{itemize}

\section{Background and Related Work}
\label{section::background and related work}
\subsection{Federated Learning}
The whole population $D = \cup_{k=1}^N D_k$ is allocated to $N$ clients, and each client maintains $D_k$. Each client maintains a local model trained from the local training dataset. And a central server maintains a global model by aggregating the local model updates from the participating client in each training round. The objective of FL training is to minimize the loss:
\begin{align}
F(w) = \frac{1}{|D|} \sum_{(x, y) \in D} L(w; (x, y)).
\end{align}
To achieve this goal, each client $k$ optimizes their local model weights $w_k$ to minimize the loss function $F_k(w) = \frac{1}{|D_k|} \sum_{(x, y) \in D_k} L(w; (x, y))$. Here, we describe the FedAvg aggregation method \cite{mcmahan2017communication}, which is perhaps the most widely used averaging scheme. FedAvg iteratively performs the following three steps:

\textbf{(1) Global model synchronization.}
In the $T$-th aggregation, the central server randomly selects $K$ $(K \leq N)$ from $N$ clients and broadcasts the latest global model $w^{T}$ to the selected clients: $\begin{aligned} w^{T, 0}_{k} \leftarrow w^{T} \end{aligned}$. 

\textbf{(2) Local model training. }
Each client $k$ updates its own local model $w^{T}_{k}$ by running an SGD on the local dataset $D_k$ for $t$ steps. The $\tau$-th step on client $k$ follows:
\begin{align}
w^{T, \tau+1}_{k} \leftarrow w^{T, \tau}_{k} - \eta \nabla F_k(w^{T, \tau}), 
\end{align}
where $\eta$ is the local learning rate. 

\textbf{(3) Global model update.}
After performing local training for $t$ steps, the client transmits the model update $\Delta w_k^T = w_k^{T, t} - w_k^{T, 0}$ back to the central server. The central server then updates the global model by performing a weighted average on the local model updates sent from $K$ clients: 
\begin{align}
\label{eq::fedavg}
w^{T+1} \leftarrow w^T + \sum_{k=1}^{K} \frac{n_k}{n} \Delta w^{T}_{k},
\end{align}
where $n_k = |D_k|$ is the number of training data on the client $k$ and $ n = \sum_{k=1}^{K} n_k$ is the total number of training data used in the selected clients. 

\subsection{Information Leakage in FL}
We mainly discuss the literature related to property inference attack. The property inference attack was firstly proposed by Ateniese et al. \cite{ateniese2015hacking} against Hidden Markov Models and Support Vector Machines. Authors in \cite{ganju2018property} designed a property inference attack on fully connected networks, in which the adversary trains a meta-classifier to classify the target classifier depending on if it possesses the interested property or not. A malicious user can infer attributes that characterize the entire data class or a subset of data \citep{melis2019exploiting}.

We also note that our whole population distribution inference attack is similar to the one in \citep{wang2019eavesdrop}, where the authors analyzed the relationship between the number of data sample of a specific label and the magnitude of the corresponding gradients. Our work differs from their work from the following two perspectives: (1) their work draws a comparison between a pair of labels and generates a binary output of which label possesses a larger number of data samples, while our work is able to provide a precise quantitative distribution of all labels. (2) to get a satisfying inference result of the whole distribution, their work has a high computation complexity and needs to be performed multiple rounds, while in our work the distribution can be inferred in one training round and requires less computation. 

\subsection{Backdoor Attacks against FL}
The backdoor attack is one of the data poisoning attacks whose goal is to misclassify inputs with backdoor triggers as the target class, while not affecting the model accuracy on clean data. The backdoor attack was first introduced in \cite{bagdasaryan2018backdoor}. They also proposed train-and-scale and constrain-and-scale techniques to maximize the attack impact while evading the anomaly detection. The researchers in \cite{wang2020attack} introduced an edge case backdoor that targets the data on the tail of the distribution. They also claimed that the backdoors against FL are unavoidable and computationally hard to detect. To make the backdoor more stealthy, scholars in \cite{xie2019dba} decomposed a centralized backdoor into parts, and each trigger is injected by one client. The distributed backdoor is more effective and persistent than the centralized backdoor. However, the distributed backdoor is brought into full play upon the completion of the injection of all distributed triggers. Additionally, to survive the newcoming normal updates, the injection of local triggers must be finished in a relatively short attack window. Given that the attacker cannot manipulate the timing of a compromised client being selected to participate in the training, the above conditions are hardly satisfied in practice. 

\subsection{Defenses against FL Backdoor Attacks}
Defense against backdoor attacks mainly falls into two categories, robust aggregation and differential privacy.

\textbf{Robust aggregation.} One approach from existing works focus on building a robust aggregation algorithm that estimates the most possible aggregation instead of directly taking a weighted average. These robust aggregation, such as Foolsgold \citep{fung2018mitigating}, Krum \citep{blanchard2017machine}, Bulyan \citep{guerraoui2018hidden}, RFA \citep{pillutla2019robust} and trimmed mean \citep{yin2018byzantine} are designed based on the statistical characteristics of model updates, and they aim to identify and deemphasize possibly malicious model updates in the aggregation. Most of the robust aggregations are built on an assumption of the i.i.d data distribution across the participating clients. However, this assumption is hardly met in practice. For the FL with non-i.i.d data among clients, robust aggregation algorithms could mis-identify the non-i.i.d. but normal model updates as malicious or vice versa, and then their weight could be reduced or raised in the aggregation, which degrades the FL model accuracy. These approaches are capable to mitigate the impact of malicious model updates to a certain level, but cannot fully eliminate them \cite{li2020learning}. 

\textbf{Differential Privacy (DP).} DP was originally designed to protect individual privacy. Scholars in \cite{sun2019can} discovered that by adding  noises the model update could also reduce the effect of the malicious model updates. It has been proved that DP is effective in mitigating backdoor attacks, but at the cost of model utility loss. Scholars in \cite{naseri2020local} evaluated the effectiveness of both local DP and central DP in defending against backdoor attacks. 

\section{Threat Model and Attack Design Philosophy}
\label{section::Threat Model and Attack Design Philosophy}
\subsection{Threat Model}
We consider the single-shot attack scenario, in which the attacker-controlled client has only one chance to inject the backdoor. And we aim to improve the effectiveness and lifespan of the backdoor injected in the early training stage. First, the attack should be kept stealthy, i.e., the impact on the main task accuracy should be as small as possible. Second, the backdoor injected in the early training stage should retain for a long period.

We assume that the attacker is able to compromise one or multiple clients, and they can interact with the FL model for multiple times. In addition to the attacker's capabilities mentioned in \cite{bagdasaryan2018backdoor}, such as local data poisoning, local training process control, we also assume that the attacker has the capability of local label distribution adjustment, in which the attacker could use data augmentation and sampling techniques to change the number of samples in each label. The ability to adjust the label distribution may vary for different attackers. To augment the data, attackers can obtain extra data samples from public datasets or use trivial techniques, such as random rotation, random zoom, random crop, etc. For attackers with strong capabilities, they can synthesize data samples from the current local dataset, and reconstruct data samples from the local dataset and gradient leakage \citep{hitaj2017deep}. This assumption is practical, as the attacker can easily integrate the above operations into data prepossessing. It is also assumed that attackers can set their own learning rate and local steps to maximize backdoor performance while minimizing the impact on the main learning task.

\subsection{Attack Design Philosophy}
Let $w_a$ denote the malicious client's local model. The single-shot backdoor attack achieves its malicious goal by attempting to substitute the new global model $w^T$ with a backdoored local model $w_a^T$ in Eq. \ref{eq::fedavg}. FL aggregation with a backdoored model update is as follows:
\begin{align}
    w^{T+1} \leftarrow w^T +  \sum_{k\neq a} \frac{n_k}{n} \Delta w_k^T + \frac{n_a}{n} \Delta w_a.
\end{align}
The malicious model $w_a$ can only fully substitute the global model by scaling up to $\gamma = \frac{n}{n_a}$ and when the global model converges, i.e., $ \sum_{k\neq a} \frac{n_k}{n} \Delta w_k^T  \approx 0$. When the FL global model converges, the newcoming clients' model updates are too small to overwrite the backdoor effect. As a result, the injected backdoor can last for a long period.

However, for the early-injected backdoor, the global model substitution is diluted due to $ \sum_{k\neq a} \frac{n_k}{n} \Delta w_k^T  \neq 0$. Combined with the effect of the scaling operation, not only the backdoor cannot reach its maximum effectiveness, but also the accuracy of the main task might deteriorate. In general, the magnitude of the model update decreases as the FL model is close to convergence. The backdoor effect is more likely to be undermined and overwritten more quickly by newcoming model updates when injected in an earlier training stage. Our main insight is that \textbf{the early-injected backdoor effect would be more effective if the early training stage convergence is expedited}, i.e., $\sum_{k\neq a} \frac{n_k}{n} \Delta w_k^T$ is reduced. 

We consider a $C$-class classification FL problem with cross-entropy as the loss function. The loss function of a client $k$ computed on its local dataset $D_k$ is defined as:
\begin{align}
 F(w; D_k) = \sum^{C}_{c=1}p_{k}(y=c)\mathbb{E}_{x\in D_k|y=c}[\log f_c(x; w_k)],
\end{align}
where $p_{k}(y=c)$ denotes the proportion of class $c$ in $D_k$, and $f_c$ is the probability of a training sample $x$ belonging to the $c$-th class.

The CL on the whole population serves as the upper bound of the FL. Due to the non-i.i.d. data distribution among participating clients, and multiple SGDs are performed on the same local dataset, the locally trained model in the FL scheme could introduce weight divergence, which deteriorates the FL global model. And this contributes to the performance gap between CL and FL. Thus, the weight divergence between the models in the CL and FL settings can be used to characterize how good an FL model is. 

Consider three models here, the local model $w_k$ of the $k$-th client, the FL global model $w$, and the CL model $w_{cen}$ trained on $D$. Previous works \cite{zhao2018federated, xiong2021privacy} have analyzed the weight divergence of the FL model $w$ and the CL model $w_{cen}$ throughout the training process and tried to catch what causes such a weight divergence. They proved that the weight divergence between $w$ and $w_{cen}$ throughout $T$ global aggregations is bounded by two terms: (1) the sum of the distribution distance between each client's local data and the whole population; (2) the weight divergence inherited from $(T-1)$-th aggregation. And such a divergence is accumulated over time, and finally leads to a model accuracy degradation.

Inspired by their work, we are more interested in the intra-aggregation weight divergence, i.e., the weight divergence between two aggregations between $w_{cen}$ and $w$, and $w_{cen}$ and $w_k$. To get rid of the influence from previous aggregations, we let the CL model and client's local synchronize with the $T$-th FL global model, $w_{cen}^{T, 0} \leftarrow w^{T}$ and $w_k^{T,0} \leftarrow w^T$. And the CL model and client's local perform $t$ steps training on the whole population data, and their weights after $\tau$ steps are:
\begin{align}
    w_{cen}^{T, \tau} &= w_{cen}^{T, \tau-1} - \eta \nabla F(w_{cen}^{T, \tau-1}; D) \nonumber \\
    &= w_{cen}^{T, \tau-1} - \eta \sum_{c=1}^{C} p(y=c) \nabla \mathbb{E}_{x \in D|y=c}[\log f_c(x; w^{T, \tau-1}_{cen})].
\end{align}

\begin{align}
    w_{k}^{T, \tau} &= w_{k}^{T, \tau-1} - \eta \nabla F(w_{k}^{T, \tau-1}; D_k) \nonumber \\
    &= w_{k}^{T, \tau-1} - \eta \sum_{c=1}^{C} p(y=c) \nabla \mathbb{E}_{x \in D_k|y=c}[\log f_c(x; w^{T, \tau-1}_{k})].
\end{align}
The weight divergence relationship among the three models can be visualized in Figure \ref{figure::weight_divergence}. We have the following proposition:
\begin{figure}[ht]
\centering
\includegraphics[width=0.6\linewidth]{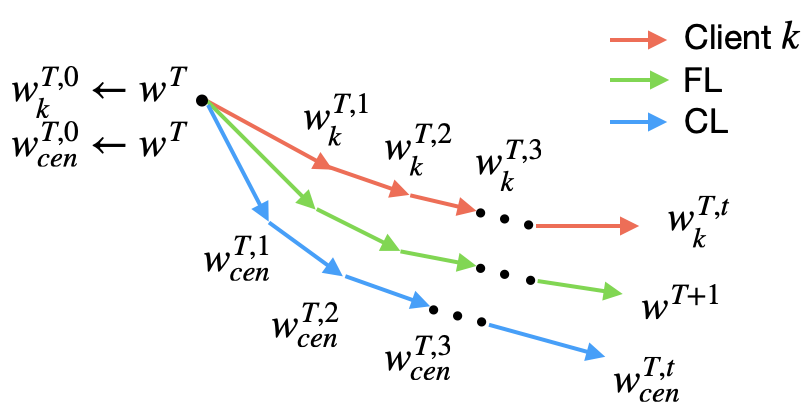}
\caption{Illustration of weight divergence relationship among an FL client's local model, FL global model, and CL model.}
\label{figure::weight_divergence}
\end{figure}

\begin{proposition}
\label{proposition}
At the $T$-th FL global aggregation, let the local model $w_{k}$ and the CL model on the entire population $w_{cen}$ synchronize with the FL global model $w^{T}$, i.e., $w_{k}^{T,0} \leftarrow w^{T}$, and $ w_{cen}^{T, 0} \leftarrow w^{T}$. And we have $p(y=c) =\sum_{k=1}^K p_{k}(y=c)$, where $p(y=c)$ and $p_k(y=c)$ are denoted as the proportion of the label $c$ on $D$ and $D_k$. Let each model train for $t$ steps, in which the global aggregation conducts. The model weight divergence between $w$ and $w_{cen}$, and $w_k$ and $w_{cen}$ after $t$ training steps are bounded by the following two equations, respectively:

\begin{small}
\begin{flalign}
\label{EQ_proposition_1}
&\|w^{T, t} - w_{cen}^{T, t}\|  \nonumber \\
\leq &  \eta \sum_{\tau=1}^t  \Big[ \big \| \sum_{c=1}^{C}\sum_{k=1}^{K} \frac{n_k}{n}  p_k(y=c)  \big[ \nabla \mathbb{E}_{x \in D_k|y=c}[\log(f_c(w^{T, \tau-1}_{k})]  - \nabla \mathbb{E}_{x \in D|y=c}[\log(f_c(w_{cen}^{T, \tau-1})] \big] \big\| \Big] 
\end{flalign}
\end{small}

\begin{small}
\begin{flalign}
\label{EQ_proposition_2}
&\|w_{k}^{T, t} - w_{cen}^{T, t}\| \nonumber \\
\leq &  \eta \sum_{\tau=1}^t   \Big[  \big \| \sum_{c=1}^{C} \big[(p(y=c) -p_{k}(y=c)\big] \nabla \mathbb{E}_{x \in D|y=c}[\log(f_c(w^{T, \tau-1}_{k})] \big\| \nonumber \\
& + \big \| \sum_{c=1}^{C} p_k(y=c)\big[\nabla \mathbb{E}_{x \in D_k|y=c}[\log f_c(x; w^{T, \tau-1}_{k})]-\nabla \mathbb{E}_{x \in D|y=c}[\log(f_c(x; w^{T, \tau-1}_{cen})]\big] \big\| \Big] 
\end{flalign}
\end{small}
\end{proposition}

The proof can be found in \ref{section::appendix}, and we have the following remarks.

\begin{remark}
\label{remark_1}
The intra-aggregation weight divergence $\|w^{T} - w_{cen}^{T}\|$ is determined by the distance between the gradient of the local model taken on $D_k, k \in [1, K]$ and the gradient of the CL model taken on $D$. This gradient distance can be reduced by increasing the local data sample size. The weight divergence is also an increasing function of internal training steps $t$. Therefore, increasing the number of local data samples or decreasing the internal training steps could mitigate weight divergence. 
\end{remark}

\begin{remark}
\label{remark_2}
The intra-aggregation weight divergence $\|w_k^{T} - w_{cen}^{T}\|$ is mainly due to two parts, which are the distribution distance between $D_k$ and $D$, that is, $\sum_{c=1}^{C}\big(p_{k}(y=c) - p(y=c)\big)$, and the gradient distance between the gradient calculated on $D_k$ and the gradient calculated on $D$ over classes, that is, $ \big[\nabla \mathbb{E}_{x \in D_k|y=c}[\log f_c(x; w^{T, t-1}_{k})]-\nabla \mathbb{E}_{x \in D|y=c}[\log(f_c(x; w^{T, t-1}_{cen})]\big]$.
\end{remark}

According to Remark \ref{remark_2}, the weight divergence $\|w^{T, t} - w_{cen}^{T, t}\|$ could be mitigated by reducing the following two terms:
(1) the difference between the data distribution of $D_k$ and that of $D$, implying the first term in the Eq. \ref{EQ_proposition_2} is reduced; (2) the difference between the gradient calculated on $D_k$ and that calculated on $D$, implying the second term in the Eq. \ref{EQ_proposition_2} is reduced.

As a result, a client in FL setting could benefit from mimicking the distribution and gradients of the whole population to achieve a better convergence behavior (faster convergence or higher model accuracy). The above finding is a double-edged sword. On the one hand, a benign client can use it to alleviate weight divergence so as to facilitate FL convergence, as the data sharing strategy proposed in \cite{zhao2018federated}. On the other hand, the finding could also be taken advantage by an adversary. As will be shown in the next section, we propose a two-phase backdoor attack, in which the above finding is utilized by an adversary to improve the FL global convergence performance, and further enhance both the strength and persistence for the subsequent single-shot backdoor injection. 

\section{Our Approach}
\label{section::our_approach}
In this section, leveraging the aforementioned insights, we present an overview of our proposed two-phase backdoor attack. Then we describe the detailed workflow of the proposed backdoor attack. 

\subsection{Overview}
\begin{figure}[ht!]
\centering
\includegraphics[width=0.8\linewidth]{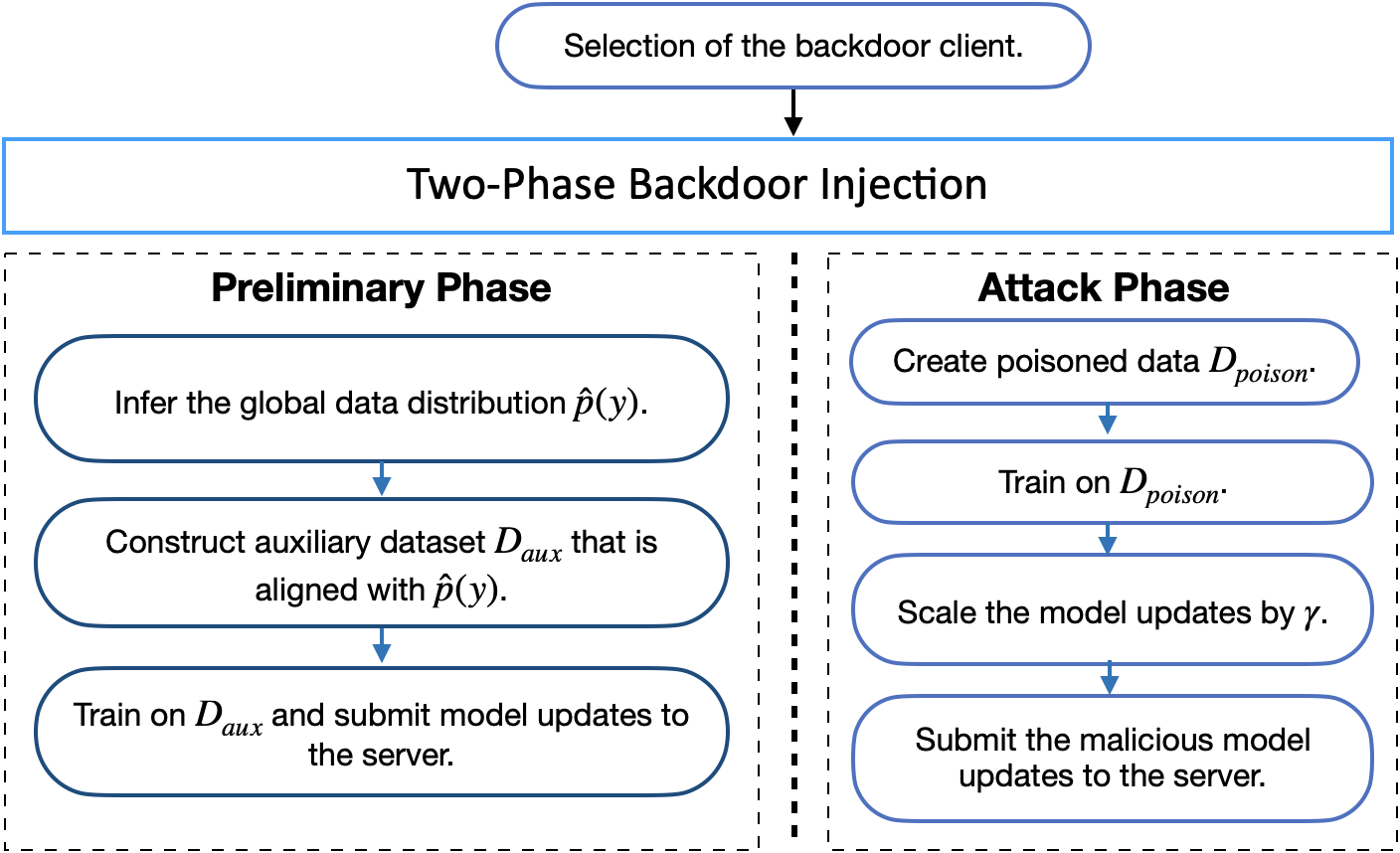}
\caption{The flow chart of the proposed two-phase backdoor attack.}
\label{figure::flow_chart}
\end{figure}
Our proposed two-phase backdoor attack, illustrated in Figure \ref{figure::flow_chart}, consists of a preliminary phase and an attack phase. The backdoor attack can be any kind of existing backdoor attack. Our approach is different from the existing backdoor attacks in the proposed preliminary phase ahead of the attack. The goal of the preliminary phase is to expedite the FL model convergence such that the subsequent backdoor can be more effective and consistent. Specifically, the attacker-controlled client first launches a passive whole population distribution inference attack by analyzing their local model updates and the FL global model update. To reduce the weight divergence and improve the convergence behavior of the FL model, attacker-compromised client then crafts the local training data by augmentation and downsampling such that the distribution $p_k(y)$ is align with the inferred whole population distribution $\hat{p}(y)$. This step reduces the first term in Eq. \ref{EQ_proposition_2}, i.e., the distribution difference $\sum_{c=1}^C (p_k(y=c) - p(y=c))$. A dynamic sample size determination method is also utilized in the dataset crafting in order to reduce the second term in Eq. \ref{EQ_proposition_2}, i.e., the gradient distance $ \big[\nabla \mathbb{E}_{x \in D_k|y=c}[\log f_c(x; w^{T, t-1}_{k})]-\nabla \mathbb{E}_{x \in D|y=c}[\log(f_c(x; w^{T, t-1}_{cen})]\big]$. Instead of training on the original local dataset, attacker-compromised clients train on the crafted datasets and submit the model updates to the central server. These steps seem legitimate, but they benefit the subsequent injected backdoor by reducing the dilution effect from other clients' model updates. When the backdoor client is selected or the expected accuracy is reached, the adversary injects the backdoor by training on a poisoned local dataset and scales the malicious model updates by $\gamma$ to ensure the injected backdoor survives the aggregation before being submitted to the central server. 

Our proposed two-phase backdoor attack improves the performance of the early-injected backdoor because of the following features:
\begin{itemize}
    \item We propose a passive whole population distribution inference attack that requires no access to other clients' local data samples nor their model updates.
    \item By crafting the local dataset, utilizing the inferred whole population distribution and sampling techniques, we are able to reduce the FL model weight divergence, which facilitates the FL model convergence.
    \item By improving the convergence of the FL model, the backdoor global model is less diluted by model updates from other clients, leading to a stronger and longer-lasting backdoor effect. 
\end{itemize}

\subsection{Attack Workflow}
\subsubsection{Preliminary phase: whole population distribution inference}
\textbf{Step 1. Approximation of the CL model udpates. } The attacker's goal is to estimate the whole population distribution $p(y)$ in the 
following expression of the CL loss function gradient:
\begin{align}
   \nabla F(w_{cen}; D)=  \sum^{C}_{c=1}p(y=c)\nabla E_{x \in D|y=c}[\log f_c(x; w_{cen})].
   \label{EQ_cen_decompose}
\end{align}

Therefore, $p(y)$ can be calculated if the values of $\nabla E_{x \in D|y=c}[\log f_c(x; w_{cen})]$ and $\nabla F(w_{cen}; D)$ are known. Based on the findings in Remark \ref{remark_1}, we approximate the CL model update by the FL model update:
\begin{align}
   \sum_{k=1}^{K} \frac{n_k}{n} \Delta w_k \approx  \Delta w_{cen} = \eta\sum_{\tau=1}^t  \nabla F(w_{cen}^{\tau-1}; D).
   \label{EQ_substitution}
\end{align}

The reasonability of the approximation is demonstrated by: \textbf{(1) the bounded and small intra-aggregation weight divergence between the CL model and the FL model.} In Proposition \ref{proposition}, we show that the intra-aggregation weight divergence between a CL model and a FL model is bounded by the difference in the gradient of the local data and the whole population. Such a gradient difference is usually brought about by the difference in the number of samples between the local data and the whole population. The adversary could refer to public dataset or use augmentation techniques to get a good estimate of the gradient of the whole population. In addition, although the number of internal training epochs increases the bound, the number of local training epochs in practice is relatively small, usually between 2 to 5, and therefore their impact should be minor. As a result, the FL model would not deviate much from the CL model in one aggregation; \textbf{(2) the accurate global distribution inferred from the approximation.} Extensive experiments are conducted in Section \ref{subsection::inference_result} to verify that the approximation produces accurate whole distribution inference results. The settings of these experiments are comprehensive as they cover both the balanced/imbalanced global distribution and the different non-i.i.d.-ness among local data. The results under all settings show that the difference of the true and the global distribution that is inferred from the approximation is condensed and small. 

\textbf{Step 2. Decomposition of the model updates.} Combining Eq. \ref{EQ_cen_decompose} and Eq. \ref{EQ_substitution}, we have the following gradient expression:
\begin{align}
   \sum_{k=1}^{K} \frac{n_k}{n} \Delta w_k \approx \Delta w_{cen}= \eta \sum_{\tau=1}^t \sum^{C}_{c=1}p(y=c)\nabla E_{x \in D|y=c}[\log f_c(x; w_{cen}^{\tau-1})].
   \label{eq::CL_update}
\end{align}

The model update of the compromised client $a$ can be expressed as:
\begin{align}
\Delta w_{a} = \eta \sum_{\tau=1}^t \nabla F_k(w_a^{\tau-1}; D_a) = \eta \sum_{\tau=1}^t \sum^{C}_{c=1}p_a(y=c)\nabla E_{x \in D_a |y=c}[\log f_c(x; w_a^{\tau-1})].
\label{eq::attacker_update}
\end{align}

Normally, the gradient is directly calculated by the partial derivative of the loss, e.g., $\nabla F_k(w_a; D_a)= \frac{\partial F_k(w_a, D_a)}{\partial w_a}$. Making use of the linearity of the cross-entropy loss, the gradient $\nabla F_k(w_{a}, D_a)$ can also be viewed as an weighted average over $\nabla E_{x \in D_a |y=c}[log f_c(x; w_a)]$. If the adversary gets a good estimate of $\nabla E_{x \in D|y=c}[\log f_c(x; w_{cen})]$, the global distribution $p(y)$ can be estimated by minimizing the difference between Eq. \ref{eq::CL_update} and Eq. \ref{eq::attacker_update}. 
 
\textbf{Step 3. Estimation of the gradients.} The difference between the gradients calculated on $D$ and $D_a$ are mainly caused by the difference in data sample sizes. Commonly, a larger size of data samples would provide a less biased estimate. The adversary could obtain a more accurate estimate of $\nabla E_{x \in D|y=c}[\log f_c(x; w_{cen})]$ by enlarging $D_a$ using public dataset or data augmentation techniques. However, purely pursuing a large data sample size is not always practical and effective, as some data augmentation methods are computationally expensive and time consuming, while others could generate similar samples, which could harm the estimation accuracy. Therefore, we adopt a dynamic data size determination algorithm proposed in \cite{byrd2012sample} to determine when to stop the augmentation. The method evaluates the amount of augmentation by measuring the directional distance between the gradient of the augmentation and the gradient estimate. A scaler $\theta \in [0, 1]$, indicating the cosine similarity between the gradient of augmentation and the gradient estimate, is used to determine when to stop the augmentation. A greater $\theta$ indicates a more accurate estimate, meanwhile a greater amount of augmentation.

\textbf{Step 4. Optimization-based global distribution estimation.} In the previous step, the attacker gets a good estimate of $\nabla E_{x \in D |y=c}[log f_c(x; a)]$ by augmenting $D_a$, the inference of whole population distribution $p(y)$ then can be formulated as an optimization problem, which seeks for a $ \hat{p}(y)$ that minimizes the difference of two losses in Eq. \ref{eq::CL_update} and Eq. \ref{eq::attacker_update}:
\begin{small}
\begin{align}
\label{equation_objective}
\hat{p}(y) = \mathop{\arg \min}_{p(y)} \text{ }& \big\|\sum_{k=1}^{K} \frac{n_k}{n} \Delta w_k^T - \eta \sum_{\tau=1}^t \sum_{c=1}^C p(y=c) \nabla E_{x \in D_a|y=c}[log f_c(x; w^{T, \tau-1}_{a})] \big \| \nonumber \\
s.t. \text{ }& \sum_{c=1}^C p(y=c) =1,
\end{align}
\end{small}
where $\sum_{k=1}^{K} \frac{n_k}{n} \Delta w_k^T$ is the FL global model update at the $T$-th aggregation and can be obtained by taking the difference between the synchronizations of the FL global model $(T-1)$ -th and $T$ -th. 

Since the distribution $p(y)$ is not differentiable, an evolution algorithm is used to solve the above optimization. The evolution algorithm begins with a randomly initialized population of $p(y)$, namely, the ``fathers''. Next, individuals in the fathers go through mutation and crossover operations with a certain probability to generate more diverse individuals, namely, the ``children''. Then, ``fathers'' and ``children'' are evaluated by an objective value, in which the individuals with better objective value will enter the next generation. Algorithm \ref{algorithm_inference} and Algorithm \ref{algorithm_evaluation} detail the steps to solve optimization.

\begin{algorithm}[ht!]
\caption{Whole population distribution inference by evolution algorithm}
\begin{algorithmic}[1]
\renewcommand{\algorithmicrequire}{\textbf{Input:}} 
\renewcommand{\algorithmicensure}{\textbf{Output:}}
\REQUIRE{Number of classes $C$, population size $S$.}
\ENSURE {An estimate of the whole population distribution $\hat{p}(y)$.} \\
\STATE $g=0$. 
\STATE Initialize the distribution population $\mathbf{p}_{0}$, which consists of $S$ individuals. Each individual $p_{0,s}$ satisfies $\sum_{c=1}^C p_{0,s}(y=c) =1$.\\
\STATE Compute the FL global model update $\Delta w^{T}$.
\STATE Evaluate individuals in population $\mathbf{p_{0}}$ by Algorithm \ref{algorithm_evaluation}.
\WHILE {the termination criterion is not satisfied}
\STATE $g=g+1$.
 \STATE Create population $\mathbf{q}_g$ by crossover and mutation of individuals from $\mathbf{p}_{g-1}$.\\
 \STATE Evaluate each individual in $\mathbf{p}_{g-1}$ in the children by Algorithm \ref{algorithm_evaluation}.
 \STATE Select $S$ best individuals to population $\mathbf{p}_g$ from the populations $\mathbf{p}_{g-1}$ and $\mathbf{q}_g$.
\ENDWHILE
\STATE Return the best individual in population $\mathbf{p}_g$.
\end{algorithmic}
\label{algorithm_inference}
\end{algorithm}

\begin{algorithm}[ht!]
\caption{Objective value evaluation.}
\begin{algorithmic}[1]
\renewcommand{\algorithmicrequire}{\textbf{Input:}} 
\renewcommand{\algorithmicensure}{\textbf{Output:}}
\REQUIRE{Number of classes $C$, internal training steps $t$, learning rate $\eta$, the global model update $\Delta w^{T}$, the label composition $p(y)$ .}
\ENSURE {The objective value defined in Eq. \ref{equation_objective}}.\\
\STATE The attacker synchronizes with the latest global model $w^{T, 0}_{a} \leftarrow w^{T}$.
\FOR {$\tau=1: t$}
    \FOR {$c=1:C$}
  \STATE The attacker calculates the gradient component on class $c$:\\
  $\nabla E_{x \in D_a|y=c}[\log f_c(x; w^{T, \tau-1}_{a})]$.
 \ENDFOR
  \STATE The model weight is updated by: \\
  \STATE $w^{T, \tau}_{a} = w^{T, \tau-1}_{a} - \eta \sum_{c=1}^C p(y=c) \nabla E_{x \in D_a|y=c}[\log f_c(x; w^{T, \tau-1}_{a})]$.

  \ENDFOR
\STATE Return the objective value $\|\Delta w^{T} - \Delta w^{T}_{a}\|$, where $\Delta w^{T}_{a} = w^{T, t}_{a} - w^{T, 0}_{a} $.
\end{algorithmic}
\label{algorithm_evaluation}
\end{algorithm}
\subsection{Preliminary phase: auxiliary dataset construction}
After the adversary gets the inference of the whole population distribution, instead of training on the original local dataset, the compromised client trains on an auxiliary dataset, which is crafted to align with the inferred global distribution. 

The basic idea of auxiliary dataset construction is to augment the data in classes with inadequate samples and downsample the data in classes with excessive samples based on the inferred whole population distribution. Algorithm \ref{algorithm_aux} describes the steps of auxiliary dataset construction. In particular, the attacker first determines the total size of the auxiliary dataset. The attacker then calculates the amount of data needed for each class by the size of the datset and the inferred global distribution. As for the augmentation operation, the adversary with a limited computation budget can use trivial techniques, such as random shift, random rotation, random shear, and random zoom, while a strong adversary could utilize more advanced methods, such as data synthesis and data reconstruction. For the downsample operation, it randomly samples from current data until the desired number of samples is reached. The auxiliary dataset crafted in this way mitigates both terms in Eq. \ref{EQ_proposition_2}. 

\begin{algorithm}[ht!]
\caption{Auxiliary dataset construction.}
\begin{algorithmic}[1]
\renewcommand{\algorithmicrequire}{\textbf{Input:}} 
\renewcommand{\algorithmicensure}{\textbf{Output:}}
\REQUIRE{Auxiliary dataset size $M$, the inferred data distribution $\hat{p}^(y)$, number of classes $C$, the compromised dataset $D_a$ }
\ENSURE {Auxiliary dataset $D_{aux}$.}\\
\STATE{Calculate the data size of each class $c$ by $M_c \leftarrow M \times \hat{p}(y=c)$ for $c=1,...,C$. }\\
\STATE Calculate the data size of each class $c$ of $D_a$, $\boldsymbol{|} D_{a}|c\boldsymbol{|}$, where $D_{a}|c := \{x|y: x \in D_{a}, y=c \}$.
\FOR {$c=1:C$}
 \IF {$\boldsymbol{|}D_{a}|c \boldsymbol{|} <M_c $ }
  \STATE Augment $\boldsymbol{|}D_{a}|c\boldsymbol{|}$ to $M_c$.
 \ELSE
  \STATE {Down-sample from $D_{a}|c$, such that $\boldsymbol{|}D_{a}|c\boldsymbol{|} = M_c$.}
 \ENDIF
\STATE {Auxiliary dataset $D_{aux} \leftarrow \cup_{c=1}^C D_{a}|c$.}
\ENDFOR
\STATE Shuffle dataset $D_{aux}$.
\STATE Return $D_{aux}$.
\end{algorithmic}
\label{algorithm_aux}
\end{algorithm}

\subsection{Attack phase: backdoor injection}
The attacker-compromised clients perform training on the crafted auxiliary dataset when selected in the FL training until the expected accuracy is reached, or the malicious client that is equipped with backdoor capability is selected. The backdoor client first poisons its local data $D_a$ by adding backdoor triggers to a subset of $D_a$, and changes their labels to a target one to form a poison data subset $D_{poison}$. The rest data is kept clean and is denoted as $D_{clean}$. Then the attacker performs local training on $D_{poison} \cup D_{clean}$ aiming to maximize the accuracy on both the main task and backdoor task: 
\begin{align*}
w_a^* = \mathop{\arg\min}_{w_{a}} [F_a(w_{a}; D_{clean} ) + F_a(w_{a}; D_{poison} )].
\end{align*}
After local training, the attacker scales the model updates by a parameter $\gamma = \frac{n}{n_a} \approx K$ to ensure that the updates of the backdoor model survive the aggregation and ideally replaces the global model. The attacker could also use constrain-and-scale or train-and-scale to improve its persistence and evade anomaly detection mechanisms. 

\subsection{Coordination of Multiple Attacker-Controlled Clients}
The above presentation of the attack process is based on a single attacker-controlled client, but it can be easily extended to the scenario where the attacker controls multiple clients. The whole population distribution inference attack can be performed by any one of the compromised clients. The inferred global distribution is then shared with other attacker-controlled clients, and each of them constructs and trains on the auxiliary dataset locally. The use of multiple malicious clients can further improve the accuracy of the FL model.

\section{Experimental Setup}
\label{section::experiments_setup}
\subsection{Dataset}
We evaluate our proposed method on the hand-written digit recognition dataset, the MNIST \citep{lecun2010mnist}. The dataset contains 60,000 training data samples and 10,000 testing data samples. Each data sample is a square $28\times28$ pixel image of hand-written single digit between 0 to 9. 

\subsection{Evaluation Metrics}
\begin{enumerate}
    \item \textbf{Accuracy of whole population distribution inference attack.} We measure its accuracy by the $\ell_2$ distance of the inferred whole population distribution $\hat{p}$ and the true whole population distribution $p_{global}$, i.e., $\|\hat{p} - p_{global}\|$, referred to as ``inferred-to-true''. A smaller distance indicates a more accurate inference result. And we also evaluate the $\ell_2$ distance of the original distribution on $k$-th client $p_k$ and $p_{global}$, i.e., $\|p_k - p_{global}\|$, referred to as ``original-to-true''. The difference between such two distances is positively related to the amount of weight divergence can be reduced by whole population distribution alignment.
    
    \item \textbf{Main task FL model accuracy gain by whole population distribution alignment.} We measure the FL global model accuracy as a function of training epochs for regular FL (clients train on the original datasets) and the preliminary phase assisted FL (clients train on crafted local data that is aligned with the gradients and distribution of the whole population). 
    
    \item \textbf{Main task FL model accuracy in presence of backdoor attack.} We also present the main task model accuracy when the backdoor attack is in place. As mentioned previously, the main task might deteriorate due to the scaling operation and the dilution from the normal model updates especially when that are large in early training stage. The server could discard the model updates if an unexpected leap or drop in the main task accuracy is observed.
    \item \textbf{Backdoor attack success rate and longevity.} Given a classifier $f(\cdot)$, the backdoor attack accuracy is defined as the portion of samples in backdoor samples that are predicted as the target label $y_t$ by the classifier:
    \begin{align*}
        Acc_{backdoor} = \frac{\{|x \in D_{poison}: f(x)=y_t\}|}{|D_{poison}|}.
    \end{align*}
    The test data is constructed by adding the backdoor triggers to the original test data samples. And to avoid the influence of the original data of the target label, we remove the data of the target label in the test data. We plot the backdoor success rate of 20 global epochs since the injection to assess their longevity.
    \end{enumerate}

\subsection{FL System Setting}
We implement the FL and the proposed two-phase backdoor attack using the PyTorch framework. We conduct our experiments on the Google Colab Pro (CPU: Intel(R) Xeon(R) CPU @ 2.20GHz; RAM: 13 GB; GPU: Tesla P100-PCIE-16GB with CUDA 11.2). 

The dataset is allocated to 100 clients. In each global model aggregation, 10 clients are randomly selected to participate the FL training. Each client maintains a local model consisting of 2 convolutional layers and 2 fully connected layers. We build four distributed MNIST datasets (Table 1) to cover both balanced/imbalanced whole population and different non-i.i.d.-ness among clients' local data. The global imbalance is simulated by randomly sampling $50\%-100\%$ for each class from the original dataset. And we use the Dirichlet distribution \citep{minka2000estimating} with a hyper-parameter $\alpha$ to generate different data distributions among clients, where a smaller $\alpha$ indicates a greater non-i.i.d.-ness. 

\begin{table}[ht!]
\centering
\caption{MNIST dataset settings.}
\begin{tabular}{ c c c }
\hline
 Settings & Whole population & Local distribution \\ 
 \hline
 1 & balanced & non-i.i.d., $\alpha=1$\\ 
 2 & balanced & non-i.i.d., $\alpha=0.1$\\
 3 & imbalanced & non-i.i.d., $\alpha=1$ \\
 4 & imbalanced & non-i.i.d., $\alpha=0.1$\\
 \hline
\end{tabular}
\label{settings}
\end{table} 


\subsubsection{Preliminary phase} 
The clients are randomly selected to participate in a training round, with a certain fraction of clients training on $D_{aux}$, which is aligned with the whole population by Algorithm \ref{algorithm_aux}. The FL model is trained with full-batch gradient descent with internal epoch $t=1$ and learning rate $\eta=0.1$. 

As specified in Section \ref{section::results}, the adversary has the capability of augmenting the local dataset by augmentation techniques or accessing public datasets. In our experiment, the adversary is equipped with trivial augmentation methods. We also assume that the attacker holds $1\%$ of the MNIST dataset, from which the attacker can draw data samples and complement the auxiliary dataset. 
In the dynamic data size determination algorithm that determines when to stop the augmentation, we set $\theta=0.8$, which means that the augmentation operation stops when the cosine similarity between the gradient of augmentation and the gradient estimate reaches $0.8$. To avoid the influence of the size of $D_{aux}$, we set the size of $D_{aux}$ to be the same as that of the original dataset. The fractions of clients controlled by the attacker are chosen to be $5\%$, $10\%$ and $20\%$ of the total number of clients, denoted as ``ours\_5'', ``ours\_10'' and ``ours\_20'', respectively. And they are collectively referred to as ``ours''.

\subsubsection{Attack phase} We use pixel-pattern backdoors, as the same as those in \citep{xie2019dba, bagdasaryan2018backdoor}. We set the $4\times 4$ pixels at the upper-left corner of the image to white (pixel value of 0), and swap the label with the target label "0". The ratio of the size of the backdoor trigger to the size of the data sample is $2\%$. 

The performance of the proposed backdoor (both the main task accuracy and backdoor success rate) is evaluated on an FL with mini-batch gradient descent with a batch size of $128$. The backdoor client poisons $40$ out of $128$ data samples in each mini-batch and locally trains for poison epochs of 10 with a poison learning rate of 0.05. The global learning rate is as same as the local learning rate $\eta = 0.1$. The scaling factor is $\gamma = K = 10$. 

\section{Experimental Results}
\label{section::results}
\subsection{Accuracy of the Whole Population Distribution Inference}
\label{subsection::inference_result}
\begin{figure}[ht!]
\centering
\setlength{\textfloatsep}{0pt}
\includegraphics[width=0.8 \linewidth]{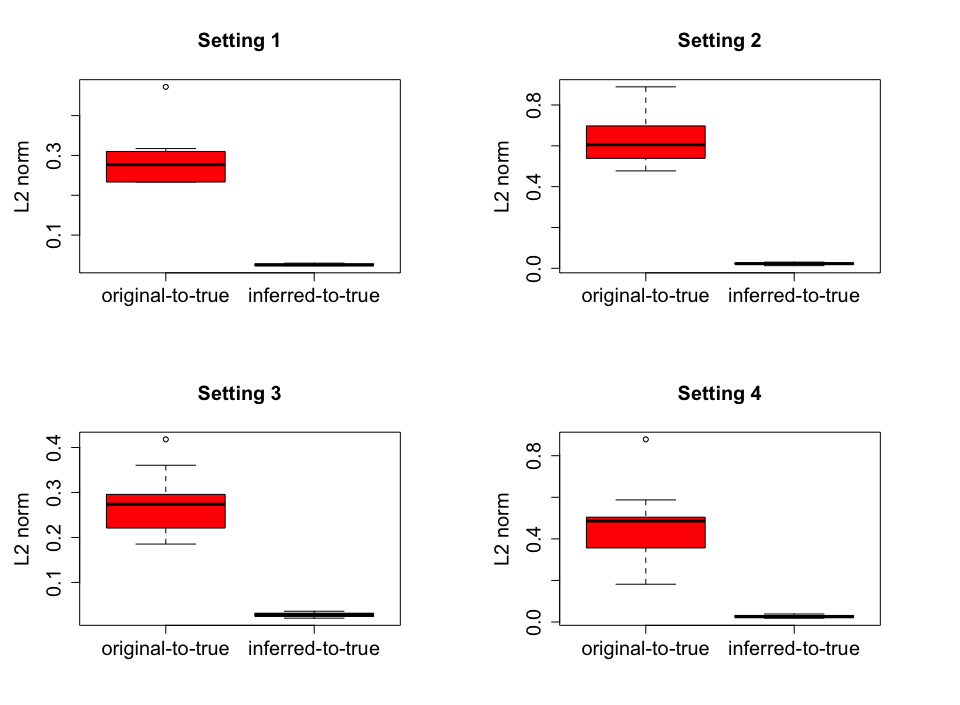}
\caption{Box plot of $\|p_k - p_{global}\|$ (``original-to-true'') and $\|\hat{p} - p_{global}\|$ (``inferred-to-true'').}
\label{boxplot}
\end{figure}

The global distribution inference attack is launched at every epoch of the first 30 epochs. We present the box plot of $\|p_k - p_{global}\|$ (referred to as ``original-to-true") and $\|\hat{p} - p_{global}\|$ (referred to as ``inferred-to-true") in Fig. \ref{boxplot}. In all four settings, compared to ``original-to-true", ``inferred-to-true" is significantly smaller and more condensed, indicating the proposed whole population distribution inference attack achieves high accuracy. Furthermore, our proposed inference attack is equally accurate in both balanced and imbalanced whole population distribution settings (setting 1 vs. setting 2 and setting 3 vs. setting 4).

\begin{figure}[ht!]
\centering
\setlength{\textfloatsep}{0pt}
\includegraphics[width=0.8 \linewidth]{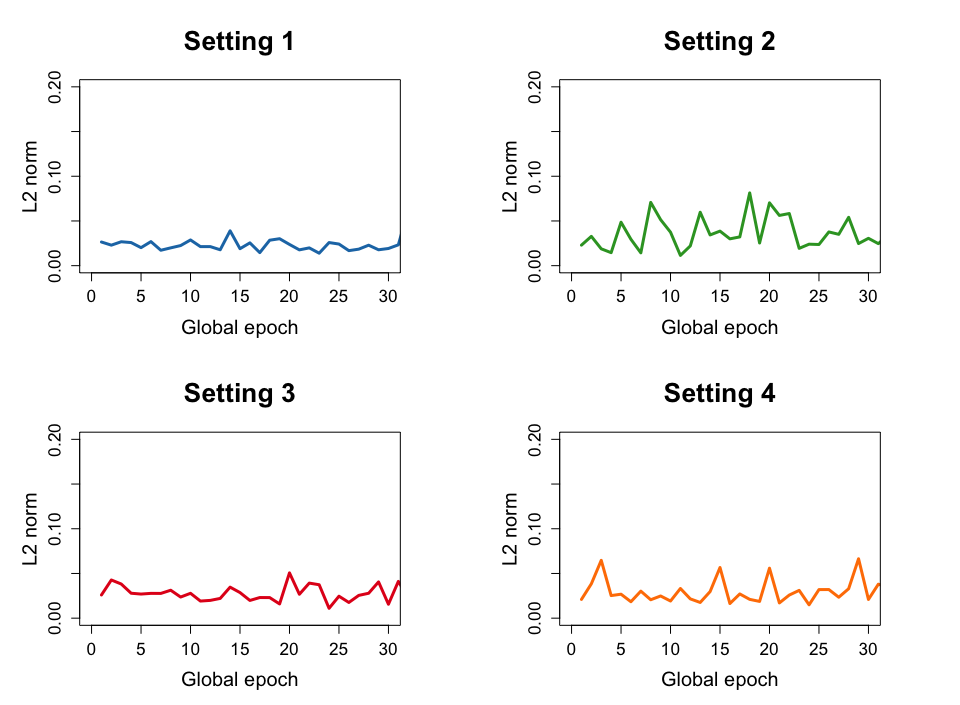}
\caption{$\|\hat{p} - p_{global}\|$ (``inferred-to-true'') vs. the global training epoch.}
\label{lineplot}
\end{figure}

We also plot the ``inferred-to-true'' as a function of training epochs (shown in Fig. \ref{lineplot}). The FL model begins to converge at epoch 20, so our inference attack window covers different convergence stages of the training process. Results show that the inference results are stationary along the training process, meaning that inferring at any training stage does not affect the inference accuracy. The fluctuations presented in Fig. \ref{lineplot} are due to the randomness of local distributions in the selected clients in each FL training round. Especially, the fluctuation becomes more noticeable when clients' local distributions are more non-i.i.d. (setting 2 and setting 4). To further reduce such fluctuations and improve the accuracy of the inference, the adversary could further refine the inference result by performing statistical analysis on multiple inference results, such as averaging or clustering.

\subsection{Main Task Accuracy under the Non-Attack Scenario}
\begin{figure}[ht!]
\centering
\includegraphics[width=\linewidth]{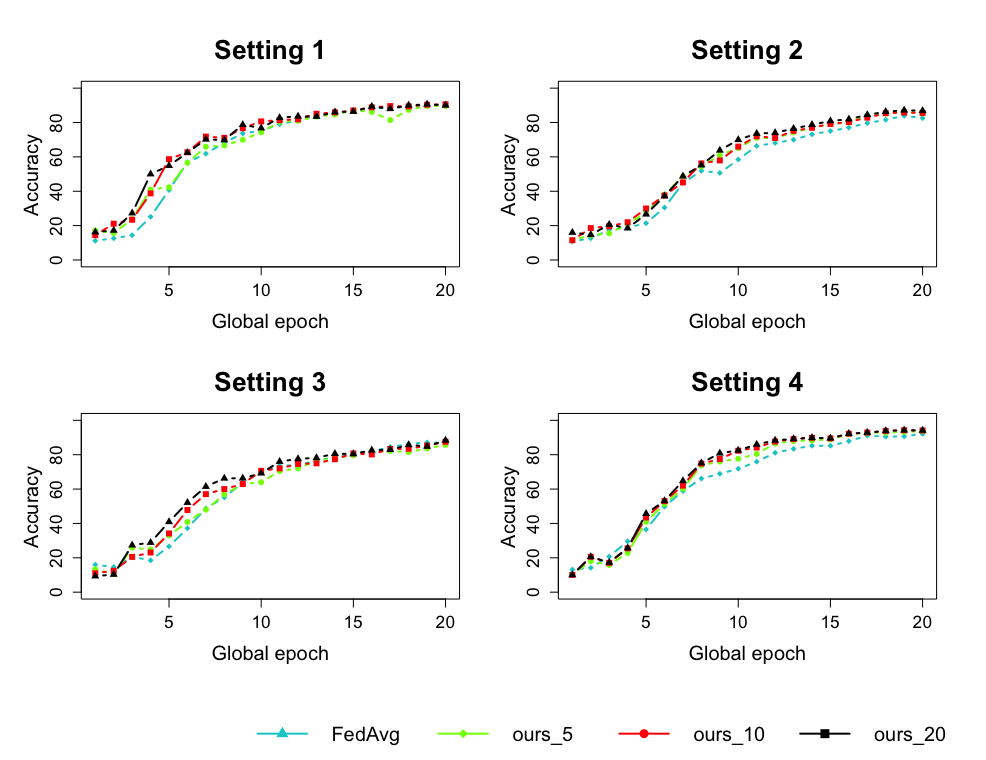}
\caption{The accuracy of the main task of $5\%$, $10\%$, and $20\%$ of the local data of the clients that perform the alignment in 4 settings, averaged over 10 experiments.}
\label{convergence}
\end{figure}

We evaluate the effectiveness of the proposed preliminary phase in improving FL convergence by the accuracy of the FL main task, shown in Fig. \ref{convergence}. Under all 4 settings, compared to FedAvg, the FL with global distribution alignment converges faster, although they eventually reach the same accuracy. Such performance gain is more perceptible before FL begins to converge and when a greater fraction of clients perform the proposed alignment. In addition, while the global distribution alignment has more influence on the very early stage (epoch 0 to epoch 10) for setting 1 and setting 3 ($\alpha=1$), a higher non-i.i.d.-ness ($\alpha=0.1$ in setting 2 and setting 4) has more impact on the middle training stage (epoch 5 to epoch 15).  The experimental results are consistent with the findings in Proposition \ref{proposition}: reducing both the gradient and distribution between the client's local data and the whole population could reduce the model weight divergence, leading to a better convergence performance.

\subsection{Backdoor Attack Performance}
We present the impact of backdoor injection on the main task accuracy as well as the backdoor success rate. We evaluate the proposed two-phase backdoor attack and compare it with two existing backdoor attacks: (1) the centralized backdoor attack \cite{bagdasaryan2018backdoor} (referred to as ``baseline''), in which the local dataset is poisoned by a centralized backdoor trigger; (2) the distributed backdoor attack \citep{xie2019dba} (referred to as ``DBA''), in which the backdoor trigger is partitioned into parts and each part is injected separately. 

\subsubsection{Main task accuracy }
Unlike the backdoors injected at the convergence of the FL model, where the injection of the backdoor barely disturbs the accuracy of the main task, the early-injected backdoor usually noticeably deteriorates the accuracy of the main task due to the not small enough model updates from normal clients. When the backdoor is injected in the early training stage, the accuracy of the main task usually experiences a sudden drop and then gradually goes back to normal status afterward. As introduced in \cite{shayan2018biscotti}, the central server could monitor the FL model main task accuracy and reject model updates that make the main task accuracy abnormally low. This approach could fail to be deployed on the FL system, since the central server does not always have access to the model updates and test data, thus cannot measure their accuracy, or a false alarm could be triggered due to the extremely low local accuracy caused by the participation of clients with highly imbalanced local data. However, the main task accuracy can be still used to evaluate the stealthiness of the backdoor attack. 

\begin{figure}[ht!]
\centering
\begin{subfigure}[Setting 1 and setting 2.]
{\includegraphics[width=\textwidth]{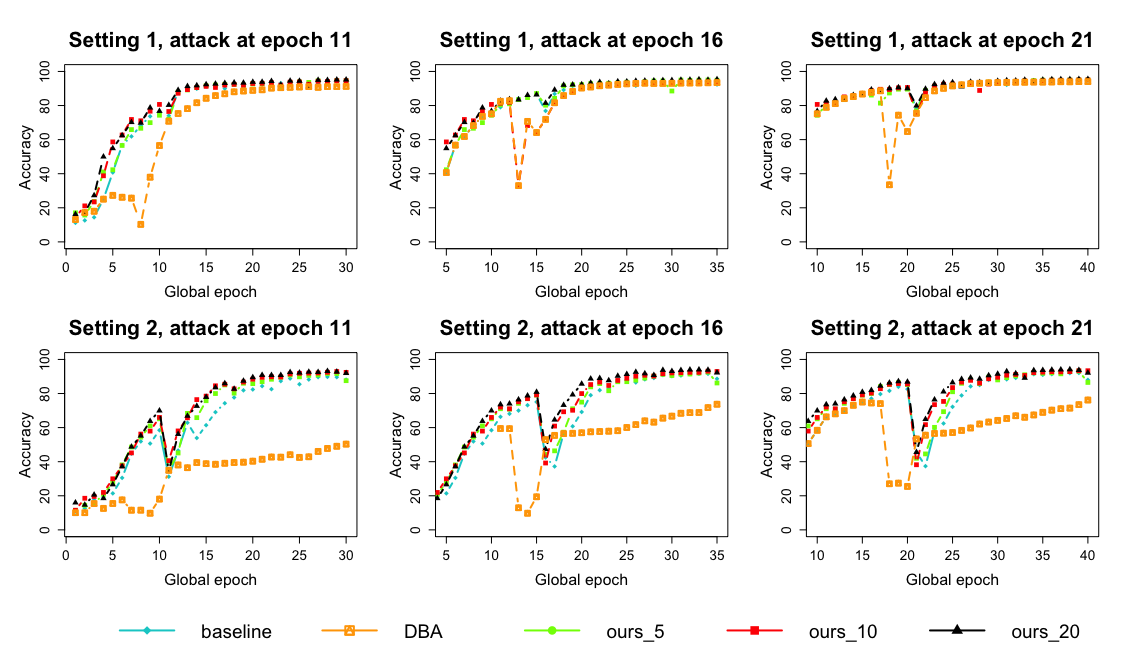}
}
\end{subfigure}
\begin{subfigure}[Setting 3 and setting 4.]{
\includegraphics[width=\textwidth]{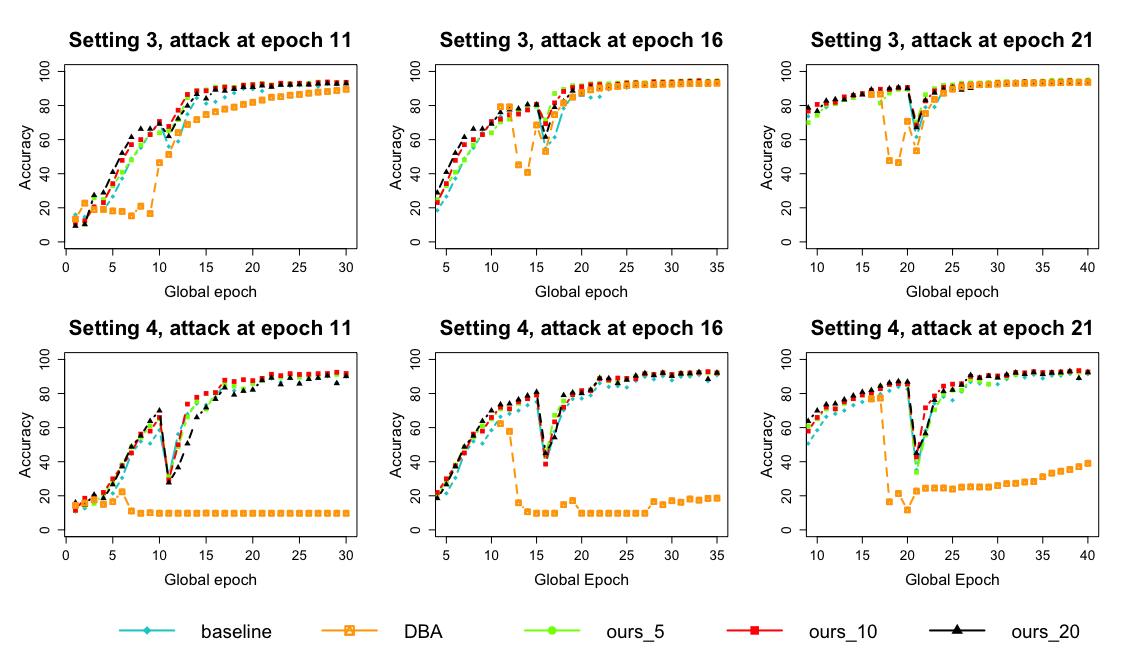}
}
\end{subfigure}
\caption{The main task accuracy of the FL global model when the backdoors are injected at FL epochs 10, 15, and 20, respectively.}
\label{fig::main_task_acc}
\end{figure}

As shown in Fig. \ref{fig::main_task_acc}, the accuracy of the main task is affected by the backdoor injection in varying degrees. The dropped main task is a collective consequence of the scaled backdoored model updates and not small enough model updates from the rest of participating clients. And such a main task accuracy drop becomes more critical for a greater non-i.i.d.-ness among clients (setting 2 and setting 4). Compared to the ``baseline'', ``ours'' introduces less drop in main task accuracy in most cases. And in some cases, the main task accuracy impacted by our proposed backdoor attack presents a faster recovery rate. Furthermore, compared to the ``baseline'' and ``ours'', the ``DBA'' suffers the greatest drop in the main task accuracy and it takes much longer for the underlying FL to return to the normal main task accuracy. This phenomenon is even worsened in the setting of high non-i.i.d.-ness (setting 2 and setting 4). A possible explanation is that ``DBA” requires multiple clients sequentially performing the injection of part of the backdoor trigger to finish the injection of a complete backdoor, which poses a longer and worse impact on the main task accuracy. Especially in the highly non-i.i.d. and globally unbalanced scenario, given that the model updates are already far from others, the consecutive injection and scale operations could make the deviation even worse and  prevent the FL model from convergence (evidenced in setting 4). Thus, we conclude that the proposed backdoor attack is more stealthy than the ``baseline'' and ``DBA''. 

\subsubsection{Backdoor attack accuracy} 
To explore the effectiveness of a backdoor before FL model converges, we inject the centralized backdoors ( ``baseline'' and ``ours'') at FL global epochs 11, 16, and 21, respectively. To fairly compare with ``DBA'', the distributed backdoors are sequentially injected and finished in the same round as the centralized backdoors. For example, if the centralized backdoor is injected in round 11, 4 distributed backdoor triggers are injected separately in rounds 8, 9, 10, and 11. 

\begin{figure}[ht!]
\centering
\subfigure[Settings 1 and setting 2]{
\includegraphics[width=\textwidth]{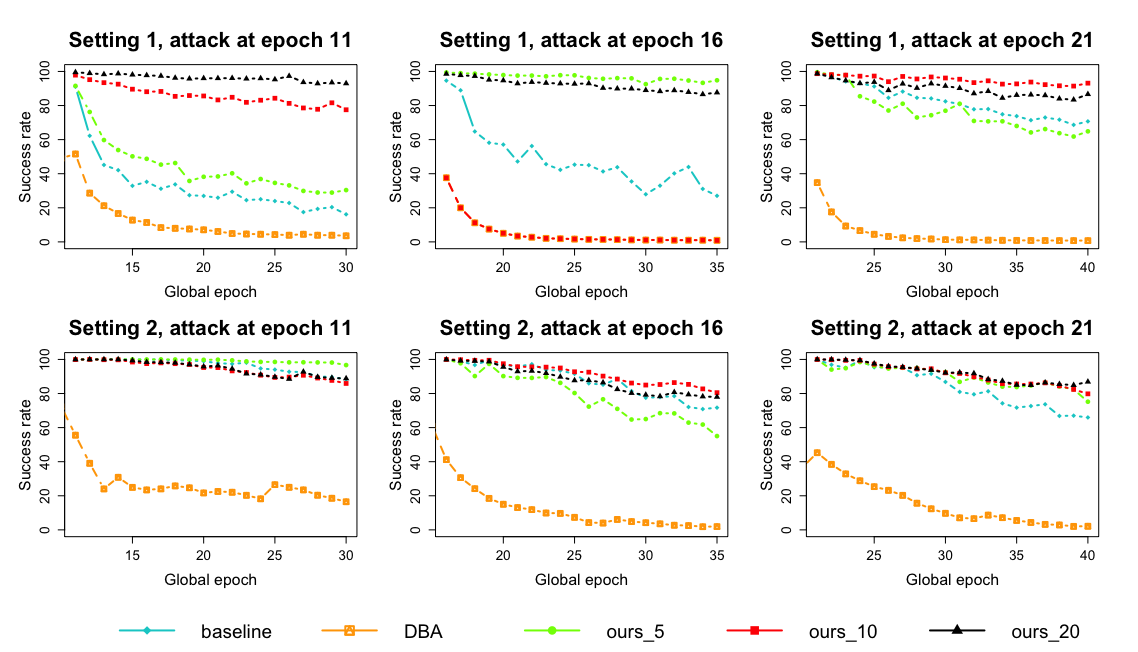}
}
\subfigure[Settings 3 and setting 4]{
\includegraphics[width=\textwidth]{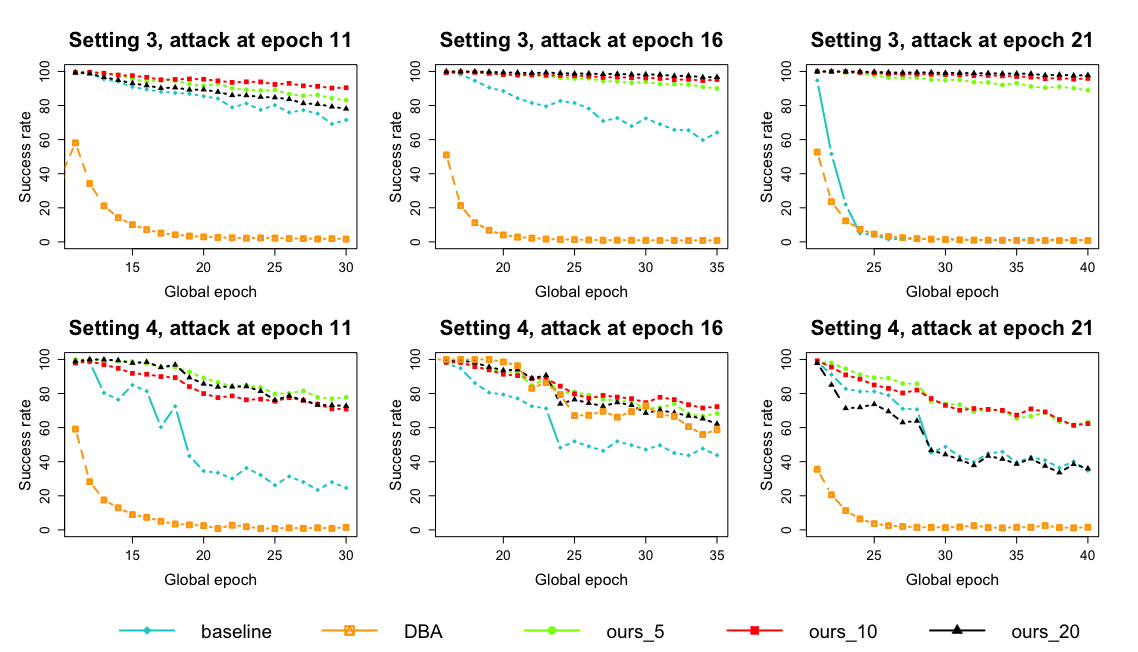}
}
\setlength{\floatsep}{-8pt}
\setlength{\intextsep}{-8pt}
\setlength{\abovecaptionskip}{-2pt}
\setlength{\belowcaptionskip}{-10pt}
\caption{The backdoor success rate in 20 training epochs since backdoor injection.}
\label{fig::backdoor_success_rate}
\end{figure}

Fig. \ref{fig::backdoor_success_rate} presents the backdoor success rate for 20 FL global epochs since the completion of the backdoor injection. For each setting, injections in different epochs are performed by the same client. The injected backdoor reached the maximum effectiveness immediately after injection. In the subsequent epochs, as the FL model aggregates new normal updates, the effect of the backdoor is weakened, which is reflected by the gradually decreasing success rate. In most of the cases, after 20 rounds since the backdoor injection, the success rates of almost all settings and injection epochs are greater than that of the ``baseline'' and ``DBA''. ``DBA'' does not reach a comparable backdoor effect as in ``baseline'' and ``ours''. The reason for this gap could be that the partially injected backdoor effect in previous rounds is more likely to be hindered by normal local updates in the subsequently injected backdoor parts. And in most cases, our proposed backdoor retains a lower diminishing rate, compared to the ``baseline''.

Due to the non-i.i.d.-ness among clients' local data, some clients' data may be in favor of the attack, while others are not. In addition, the backdoor effect does not always steadily decreases, and it bounces in some cases. Therefore, we evaluate both the attack strength and longevity  by the mean attack success rate of 10 FL epochs since injection (Table \ref{tab::backdoor_success}). In general, the backdoor injected in very early rounds (epoch 5 and epoch 10) achieves a lower mean attack success rate, compared to the ones injected in epoch 20. This degradation in the effectiveness of the attack is made even worse when the whole population is imbalanced (setting 1 vs. setting 3) and non-i.i.d.-ness among clients increases (setting 1 vs. setting 2). In most cases, our proposed backdoor attack outperforms ``baseline'' and ``DBA''. And compared to ``baseline'', the attack performance gain is positively related to the fraction of the attacker-controlled clients performing the whole population distribution alignment. 

\begin{table}[ht!]
\caption{Mean backdoor success rate(\%) over 10 FL epochs since backdoor injection (averaged over 10 randomly selected clients).}
\noindent\makebox[\textwidth]{%
\begin{tabularx}{1.2\textwidth}{@{} YYYYYY @{}}
\hline
Attack epoch& Baseline & DBA & Ours\_5  & Ours\_10 & Ours\_20\\
\hline
\multicolumn{6}{c}{\textbf{Setting 1}}\\ 
11 &$95.80\pm1.84~$& $23.25 \pm 4.19~$ & $94.06 \pm 5.97~$& $94.50 \pm 1.49~$ & $95.93 \pm 2.46~$\\
16&$97.40\pm 2.39~$& $17.23 \pm 5.36~$ & $97.40 \pm 1.34~$& $95.59 \pm 3.91~$ & $95.34 \pm 1.61~$\\
21&$95.33\pm 5.32~$& $14.46 \pm 1.32~$ & $95.32 \pm 4.24~$& $96.31 \pm 3.35~$ & $95.68 \pm 1.87~$\\
\hline
\multicolumn{6}{c}{\textbf{Setting 2}}\\ 
11&$52.91\pm 4.66$ & $77.78 \pm 28.42$ & $73.96 \pm 22.98$& $76.58 \pm 12.17$ & $80.57 \pm 27.21$\\
16&$61.25\pm 27.53$&  $67.78 \pm 10.82$ & $75.01 \pm 22.62$& $82.25 \pm 14.36$ & $78.44 \pm 17.75$\\
21&$68.38\pm 22.28$& $11.78 \pm 2.22$ & $79.24 \pm 19.67$& $79.72 \pm 13.61$ & $77.10 \pm 18.13$\\
\hline
\multicolumn{6}{c}{\textbf{Setting 3}}\\ 
11&$15.46\pm 6.35$ & $13.54 \pm 3.24$ & $44.82\pm 25.61$& $65.27 \pm 28.11$ & $66.14 \pm 25.11$\\
16&$57.71\pm 16.50$& $9.7 \pm 1.86$& $69.29\pm 18.25$& $66.11 \pm 28.66$ & $62.15 \pm 20.13$\\
21&$64.44\pm 13.77$ & $6.32 \pm 7.81$ & $73.41\pm 14.73$& $69.74 \pm 13.62$ & $72.38 \pm 16.48$\\
\hline
\multicolumn{6}{c}{\textbf{Setting 4}}\\ 
11&$48.70\pm 41.45$& $52.34 \pm 10.33$ & $67.13\pm 18.19$& $75.14\pm26.52$ & $88.28 \pm 10.11$\\
16&$68.98\pm 4.21$& $40.67 \pm 9.87$ & $72.33\pm 15.72$& $83.26\pm23.27$ & $72.22\pm 17.28$\\
21&$70.33\pm 1.94$& $11.7 \pm 4.24$ & $73.41\pm 14.73$& $88.09\pm13.39$ & $85.51 \pm 8.28$\\
\hline
\end{tabularx}}
\label{tab::backdoor_success}
\end{table}

\subsection{Overhead Analysis}
\subsubsection{Preliminary phase} 
The computational cost of this phase consists of three parts: (1) calculating the gradients on the data of each label; (2) solving the optimization in Eq. \ref{equation_objective}; (3) constructing the auxiliary dataset. 

For the first part, the attacker trains the FL global model on the data samples of each label separately to obtain gradients $\nabla \mathbb{E}_{x \in D_a |y=c} [\log f_c(x; w_a)]$, and because $n = \sum_{c=1}^C n_c$, where $n_c$ is the number of samples in label $c$, the time complexity is the same as that of local training. Since the batch gradient has a time complexity of $\mathcal{O}(n^2 m)$, in which $n$ is the number of data samples and $m$ is the number of features, the  time complexity of the first part is also $\mathcal{O}(n^2 m)$. 

\begin{figure}[ht]
\centering
\includegraphics[width=0.5\linewidth]{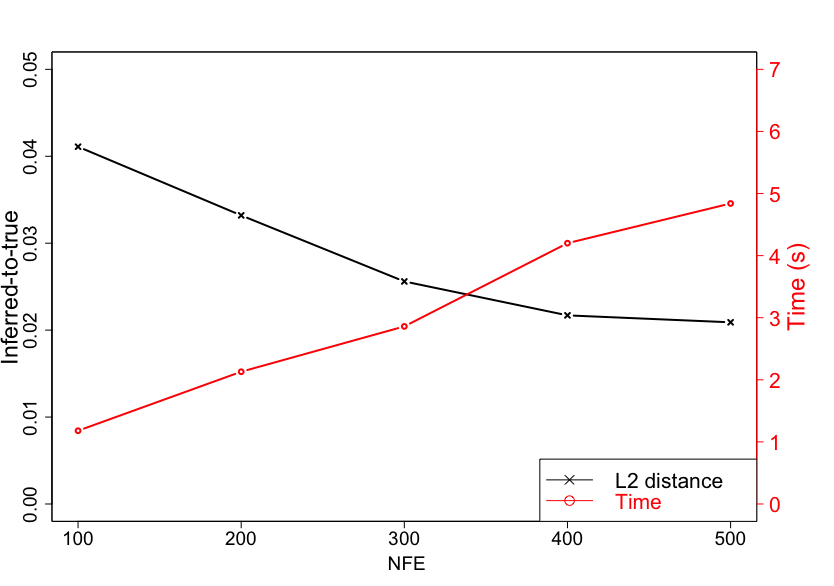}
\caption{The inference accuracy (``inferred-to-true'') and time taken vs. NFE.}
\label{fig_time_complexity}
\end{figure}

For the second part, we evaluate the \textit{number of function evaluations} (NFEs), which is commonly used to evaluate an evolution algorithm. NFE is usually measured when a good solution is delivered or when no significant change in the solution is observed. We plot ``inferred-to-true'' and the real time used against NFE, shown in Fig. \ref{fig_time_complexity}, to demonstrate the effect of NFE on inference accuracy and inference time. Because there is no significant inference accuracy gain after 400 NFEs, we set the NFE to be 400 in our experiment. The real time taken for solving the optimization with $400$ NFEs is 4 seconds.

Constructing the auxiliary dataset that is aligned with the whole population consists of augmentation and sampling operations. Examples of trivial augmentation methods are flipping ($\mathcal{O}(np)$, where $p$ is the number of pixels in each image), rotation, random crop and scale (they have the same complexity of $\mathcal{O}(n)$). The sampling operation has a time complexity of $\mathcal{O}(n)$. Therefore, the total time complexity of the construction of the auxiliary dataset is at most $\mathcal{O}(np)$ and the real time spent is $0.03$ seconds.

\subsubsection{Backdoor phase.} The backdoor client poisons a subset of the local data by injecting the backdoor pattern and swaps the label to the target ones, then performs local training on the poisoned local dataset. The total time complexity is $\mathcal{O}(n^2m)$. The real time spent on the backdoor attack with 10 internal training epochs is around 13 seconds. 

The complexity analyses are summarized in Table \ref{tab_time_complexity}. The gradient calculation and solving optimization are only needed to be performed a few times to get an accurate whole population distribution inference result. The real time for these two steps is less than $5$ seconds, which is minor compared to the time taken for the backdoor attack. Once the whole population distribution is inferred, the attacker-controlled clients only perform the auxiliary dataset construction, whose time complexity is negligible. 

\begin{table}[htbp]
    \centering
    \caption{Time complexity and real time spent on the proposed inference attack.}
\begin{tabular}{c c c}
\hline
Operation & Time complexity & Real time taken (s)\\
\hline
 Gradient calculation\  & $\mathcal{O}(n^2 m)$ & 0.11\\
Solving optimization  & 400 NFEs & 4.20 \\
 Auxiliary dataset construction & $\mathcal{O}(nk)$ & 0.03 \\
 Backdoor attack & $\mathcal{O}(n^2 m)$ & 13.37 \\
\hline
    \end{tabular}
    \label{tab_time_complexity}
\end{table}

\section{The Robustness of the Proposed Attack}
\label{section::defense}
In this section, we are interested in how the proposed attacks will behave when defense mechanisms are in place. In the following, we will analyze the effectiveness of the proposed  two-phase backdoor attack against two mainstream defense strategies. 

\textbf{FoolsGold} is a secure aggregation strategy, which calculates the cosine similarity of all historical gradient records and assigns smaller aggregation weights to clients that repeatedly contribute similar gradient updates \cite{fung2018mitigating}. 

\textbf{DP} is a noise-based method that limits the efficacy of backdoor attacks by two key steps \cite{naseri2020local}: (1) model parameters are clipped to bound the sensitivity of local model updates; (2) Gaussian noises are added to the local model updates. We consider local DP, in which each client adds noises before uploading the model updates to the server. We use the $(\epsilon, \delta)$-DP proposed in \cite{abadi2016deep} with a popular choice of $\sigma = \sqrt{2 \log \frac{1.25}{\delta}}/\epsilon$ with $\delta = 10^{-5}$ and $\epsilon=50$. And the clipping bound is set to the median of the norms of the unclipped local model updates during training. The noises are only applied to normal model updates, while the backdoor client sends the non-perturbed backdoored model updates.

\subsection{Whole Population Distribution Inference Accuracy against Defense Strategies.} 
\begin{figure}[ht!]
\centering
\includegraphics[width=\linewidth]{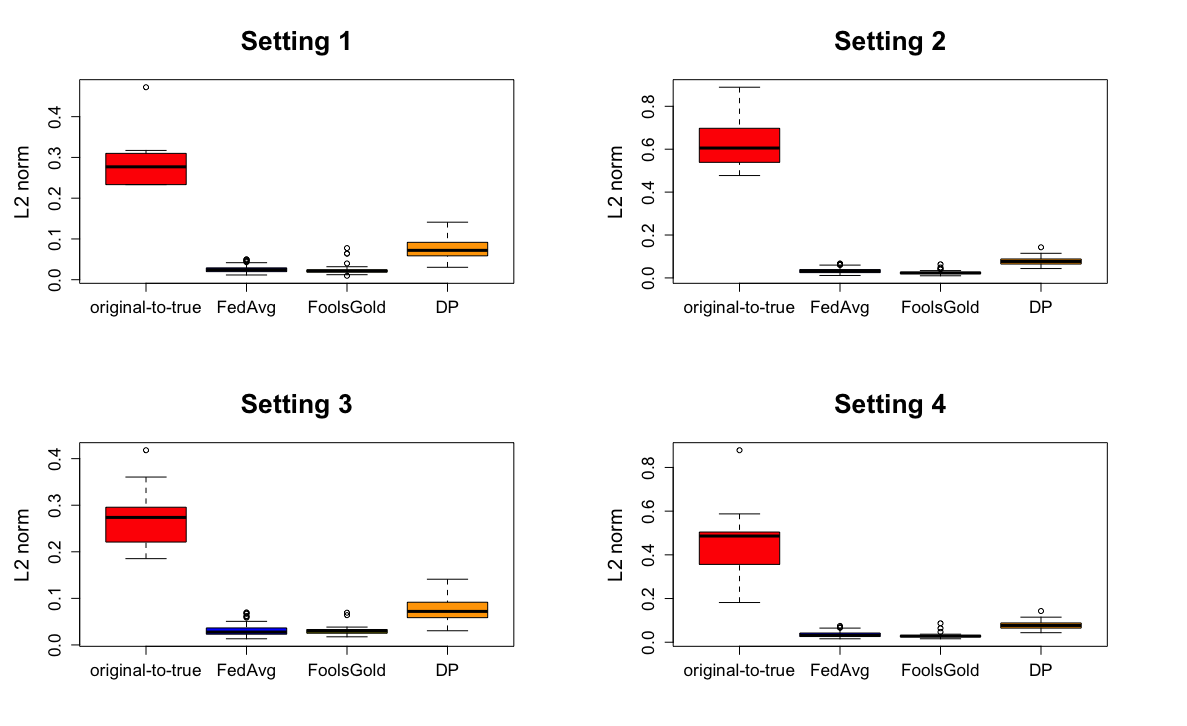}
\caption{Box plot of ``original-to-true'' and ``inferred-to-true'' ($\|\hat{p} - p_{global}\|$) of FedAvg, FoolsGold and DP based on $30$ instances.}
\label{fig::inference_accuracy_defense}
\setlength{\floatsep}{-8pt}
\setlength{\intextsep}{-8pt}
\setlength{\abovecaptionskip}{-20pt}
\setlength{\belowcaptionskip}{-10pt}
\end{figure}

We first present the whole population distribution inference accuracy against FedAvg, FoolsGold, and DP, shown in Fig. \ref{fig::inference_accuracy_defense}. Since FoolsGold does not interfere with benign FL settings, the whole population distribution inference against FoolsGold is as accurate as that in FedAvg. Although DP provides a statistical guarantee for record-level information, DP fails to protect statistical information, such as the whole population distribution. With DP in place, although the inference is not as accurate as that in the FedAvg case, the ``inferred-to-true'' is still notably lower compared with ``original-to-true''. Thus, both FoolsGold and DP fail to defend against the proposed inference attack. 

\subsection{Performance of the Backdoor Attack against Defense Strategies.} 
\begin{figure}[ht!]
\centering
\subfigure[FoolsGold]{
\label{fig::foolsgold}
\includegraphics[width=\textwidth]{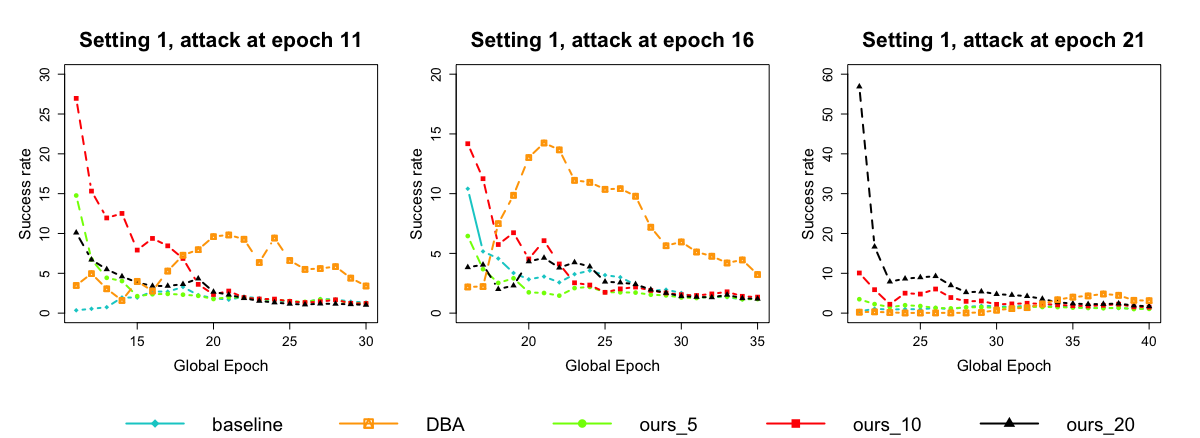}
}
\subfigure[DP]{
\includegraphics[width=\textwidth]{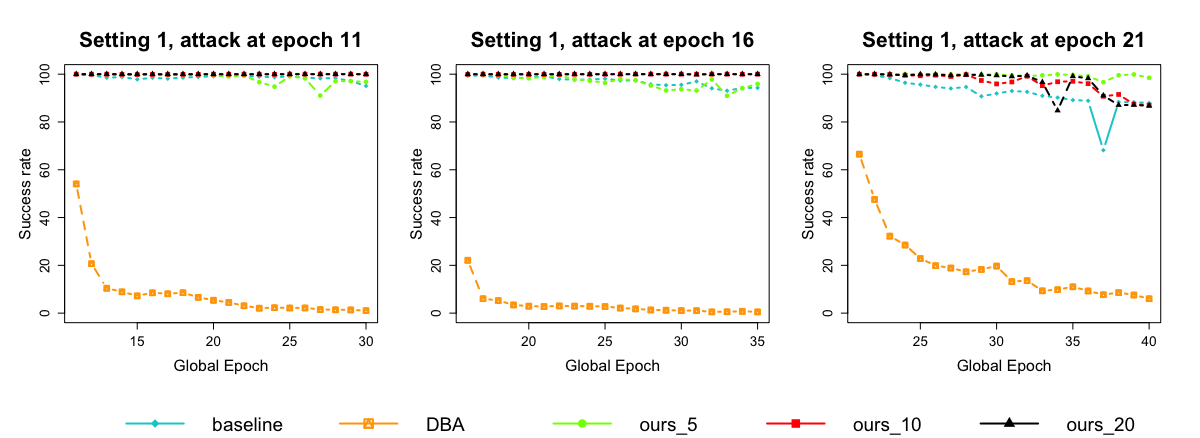}
\label{fig::DP}
}
\setlength{\floatsep}{-8pt}
\setlength{\intextsep}{-8pt}
\setlength{\abovecaptionskip}{-2pt}
\setlength{\belowcaptionskip}{-10pt}
\caption{Backdoor success rate (\%) of 20 training epochs since injection against Foolsgold (a) and DP (b) defense mechanisms.}
\label{fig::backdoor_defense}
\end{figure}

We implement the proposed backdoor attack against FoolsGold and DP in setting 1. We plot the backdoor success rate for 20 epochs after injection to observe its injection strength as well as longevity. Fig. \ref{fig::foolsgold} shows that in most cases ``ours'' reaches a significantly higher attack rate after injection through the backdoor and maintains such a high success rate in the epochs after injection. For example, when injected in the early training stage (at epoch 11), the baseline backdoor fails while ``ours\_5'', ``ours\_10''and ``ours\_20'' achieve attack success rates of $14\%$, $27\%$ and $10\%$, respectively. As the convergence is expedited by the proposed preliminary phase, the model updates from normal clients become smaller and more similar, so that the FoolsGold will reduce their assigned weights and as a result the backdoored model updates become more influential. Compared to ``ours'' and ``baseline'', the success rate curve of the ``DBA'' has a different pattern, in which the success rate first increases and then decreases and is more persistent than both the ``baseline'' and ``ours' in later rounds. However, the requirement for clients with distributed backdoor triggers to be selected in consecutive rounds can hardly be met in practice.  

Fig. \ref{fig::DP} shows the attack performance against DP. Both the ``baseline'' and ``ours'' achieve high attack success rates, even comparable to that of FedAvg, in which no defense mechanism is applied. This phenomenon indicates that instead of mitigating the backdoor effect, the noise added to the normal clients helps the backdoored model to corrupt the FL global model. A possible explanation is that the added noise reduces the utility of normal model updates, which, in turn, strengthens the backdoored model updates in the FL aggregation. In addition, ``ours'' is markedly better than the ``baseline'' when the backdoor is injected in later rounds. For example,``baseline'' and ``ours'' have similar attack performance at the early training stage, e.g., epoch 11. And ``ours'' performs distinctly better in the later training stage, that is, the backdoors injected in epoch 16 and epoch 21. Lastly, both ``baseline'' and ``ours'' outperform DBA. This is because the effectiveness of the distributed backdoor is mitigated by both benign model updates and DP noise multiple times before DBA finishes the injection of the complete backdoor.

\section{Conclusions}
\label{section::conclusion}
In this paper, we proposed a novel information leakage assisted single-shot backdoor attack that improves the effectiveness of the backdoored injected in the early training stage. We first showed that clients training on the datasets that are aligned with the whole population in both distribution and gradient can improve the FL model convergence. Based on this observation, we introduced a preliminary phase to the subsequent backdoor attack, in which the attacker-controlled clients first infer the whole population distribution from the shared FL model updates, and then they train on locally crafted datasets that are aligned with both the distribution and gradient of the whole population. Benefiting from the preliminary phase, the subsequent backdoor injection suffers less dilution effect from the model updates of other clients and achieves better effectiveness. We demonstrated the effectiveness of the proposed backdoor attacks in the early training stage through extensive experiments on a real-world dataset. Results have shown that the proposed backdoor can achieve a longer lifespan than that of the existing backdoor attacks. We hope that our work brings attention to the vulnerabilities in FL early training stage. Our analysis and findings provide novel insights for the field of strengthening FL attacks by information leakage, which could help to evaluate and improve the robustness of FL. 

\bibliography{new_bibfile}

\begin{thebibliography}{10}
\expandafter\ifx\csname url\endcsname\relax
  \def\url#1{\texttt{#1}}\fi
\expandafter\ifx\csname urlprefix\endcsname\relax\def\urlprefix{URL }\fi
\expandafter\ifx\csname href\endcsname\relax
  \def\href#1#2{#2} \def\path#1{#1}\fi

\bibitem{konevcny2016federated}
J.~Kone{\v{c}}n{\`y}, H.~B. McMahan, F.~X. Yu, P.~Richt{\'a}rik, A.~T. Suresh,
  D.~Bacon, Federated learning: Strategies for improving communication
  efficiency, arXiv preprint arXiv:1610.05492.

\bibitem{mcmahan2017communication}
B.~McMahan, E.~Moore, D.~Ramage, S.~Hampson, B.~A.~y. Arcas,
  {Communication-Efficient Learning of Deep Networks from Decentralized Data},
  in: A.~Singh, J.~Zhu (Eds.), Proceedings of the 20th International Conference
  on Artificial Intelligence and Statistics, Vol.~54 of Proceedings of Machine
  Learning Research, PMLR, 2017, pp. 1273--1282.

\bibitem{chen2019federated}
M.~Chen, R.~Mathews, T.~Ouyang, F.~Beaufays, Federated learning of
  out-of-vocabulary words, arXiv preprint arXiv:1903.10635.

\bibitem{yang2018applied}
T.~Yang, G.~Andrew, H.~Eichner, H.~Sun, W.~Li, N.~Kong, D.~Ramage, F.~Beaufays,
  Applied federated learning: Improving google keyboard query suggestions,
  arXiv preprint arXiv:1812.02903.

\bibitem{ramaswamy2019federated}
S.~Ramaswamy, R.~Mathews, K.~Rao, F.~Beaufays, Federated learning for emoji
  prediction in a mobile keyboard, arXiv preprint arXiv:1906.04329.

\bibitem{hard2018federated}
A.~Hard, K.~Rao, R.~Mathews, S.~Ramaswamy, F.~Beaufays, S.~Augenstein,
  H.~Eichner, C.~Kiddon, D.~Ramage, Federated learning for mobile keyboard
  prediction, arXiv preprint arXiv:1811.03604.

\bibitem{han2019visual}
X.~Han, H.~Yu, H.~Gu, Visual inspection with federated learning, in: F.~Karray,
  A.~Campilho, A.~Yu (Eds.), Image Analysis and Recognition, Springer
  International Publishing, Cham, 2019, pp. 52--64.

\bibitem{xu2020federated}
J.~Xu, B.~S. Glicksberg, C.~Su, P.~Walker, J.~Bian, F.~Wang, Federated learning
  for healthcare informatics, Journal of Healthcare Informatics Research 5~(1)
  (2021) 1--19.
\newblock \href {http://dx.doi.org/10.1007/s41666-020-00082-4}
  {\path{doi:10.1007/s41666-020-00082-4}}.

\bibitem{brisimi2018federated}
T.~S. Brisimi, R.~Chen, T.~Mela, A.~Olshevsky, I.~C. Paschalidis, W.~Shi,
  Federated learning of predictive models from federated electronic health
  records, International Journal of Medical Informatics 112 (2018) 59--67.
\newblock \href
  {http://dx.doi.org/https://doi.org/10.1016/j.ijmedinf.2018.01.007}
  {\path{doi:https://doi.org/10.1016/j.ijmedinf.2018.01.007}}.

\bibitem{bagdasaryan2018backdoor}
E.~Bagdasaryan, A.~Veit, Y.~Hua, D.~Estrin, V.~Shmatikov, How to backdoor
  federated learning, arXiv preprint arXiv:1807.00459.

\bibitem{wang2020attack}
H.~Wang, K.~Sreenivasan, S.~Rajput, H.~Vishwakarma, S.~Agarwal, J.-y. Sohn,
  K.~Lee, D.~Papailiopoulos, Attack of the tails: Yes, you really can backdoor
  federated learning, arXiv preprint arXiv:2007.05084.

\bibitem{dwork2010difficulties}
C.~Dwork, M.~Naor, On the difficulties of disclosure prevention in statistical
  databases or the case for differential privacy, Journal of Privacy and
  Confidentiality 2~(1).

\bibitem{fredrikson2015model}
M.~Fredrikson, S.~Jha, T.~Ristenpart, Model inversion attacks that exploit
  confidence information and basic countermeasures, in: Proceedings of the 22nd
  ACM SIGSAC Conference on Computer and Communications Security, Denver, USA,
  2015, pp. 1322--1333.
\newblock \href {http://dx.doi.org/10.1145/2810103.2813677}
  {\path{doi:10.1145/2810103.2813677}}.

\bibitem{orekondy2018gradient}
T.~Orekondy, S.~J. Oh, Y.~Zhang, B.~Schiele, M.~Fritz, Gradient-leaks:
  Understanding and controlling deanonymization in federated learning, arXiv
  preprint arXiv:1805.05838.

\bibitem{hitaj2017deep}
B.~Hitaj, G.~Ateniese, F.~Perez-Cruz, Deep models under the gan: Information
  leakage from collaborative deep learning, in: Proceedings of the 2017 ACM
  SIGSAC Conference on Computer and Communications Security, Association for
  Computing Machinery, New York, NY, USA, 2017, pp. 603--618.
\newblock \href {http://dx.doi.org/10.1145/3133956.3134012}
  {\path{doi:10.1145/3133956.3134012}}.

\bibitem{melis2019exploiting}
L.~Melis, C.~Song, E.~De~Cristofaro, V.~Shmatikov, Exploiting unintended
  feature leakage in collaborative learning, in: 2019 IEEE Symposium on
  Security and Privacy (SP), 2019, pp. 691--706.
\newblock \href {http://dx.doi.org/10.1109/SP.2019.00029}
  {\path{doi:10.1109/SP.2019.00029}}.

\bibitem{nasr2018comprehensive}
M.~Nasr, R.~Shokri, A.~Houmansadr, Comprehensive privacy analysis of deep
  learning: Stand-alone and federated learning under passive and active
  white-box inference attacks, arXiv preprint arXiv:1812.00910.

\bibitem{ateniese2015hacking}
G.~Ateniese, L.~V. Mancini, A.~Spognardi, A.~Villani, D.~Vitali, G.~Felici,
  Hacking smart machines with smarter ones: How to extract meaningful data from
  machine learning classifiers, International Journal of Security and Networks
  10~(3) (2015) 137--150.
\newblock \href {http://dx.doi.org/10.1504/IJSN.2015.071829}
  {\path{doi:10.1504/IJSN.2015.071829}}.

\bibitem{ganju2018property}
K.~Ganju, Q.~Wang, W.~Yang, C.~A. Gunter, N.~Borisov, Property inference
  attacks on fully connected neural networks using permutation invariant
  representations, in: Proceedings of the 2018 ACM SIGSAC Conference on
  Computer and Communications Security, CCS '18, Association for Computing
  Machinery, New York, NY, USA, 2018, pp. 619--633.
\newblock \href {http://dx.doi.org/10.1145/3243734.3243834}
  {\path{doi:10.1145/3243734.3243834}}.

\bibitem{wang2019eavesdrop}
L.~Wang, S.~Xu, X.~Wang, Q.~Zhu, Eavesdrop the composition proportion of
  training labels in federated learning, arXiv preprint arXiv:1910.06044.

\bibitem{xie2019dba}
C.~Xie, K.~Huang, P.-Y. Chen, B.~Li, Dba: Distributed backdoor attacks against
  federated learning, in: International Conference on Learning Representations,
  2020.

\bibitem{fung2018mitigating}
C.~Fung, C.~J. Yoon, I.~Beschastnikh, Mitigating sybils in federated learning
  poisoning, arXiv preprint arXiv:1808.04866.

\bibitem{blanchard2017machine}
P.~Blanchard, E.~M. El~Mhamdi, R.~Guerraoui, J.~Stainer, Machine learning with
  adversaries: Byzantine tolerant gradient descent, in: I.~Guyon, U.~V.
  Luxburg, S.~Bengio, H.~Wallach, R.~Fergus, S.~Vishwanathan, R.~Garnett
  (Eds.), Advances in Neural Information Processing Systems, Vol.~30, Curran
  Associates, Inc., 2017.

\bibitem{guerraoui2018hidden}
E.~M. El~Mhamdi, R.~Guerraoui, S.~Rouault, The hidden vulnerability of
  distributed learning in {B}yzantium, in: J.~Dy, A.~Krause (Eds.), Proceedings
  of the 35th International Conference on Machine Learning, Vol.~80 of
  Proceedings of Machine Learning Research, PMLR, 2018, pp. 3521--3530.

\bibitem{pillutla2019robust}
K.~Pillutla, S.~M. Kakade, Z.~Harchaoui, Robust aggregation for federated
  learning, arXiv preprint arXiv:1912.13445.

\bibitem{yin2018byzantine}
D.~Yin, Y.~Chen, R.~Kannan, P.~Bartlett, Byzantine-robust distributed learning:
  Towards optimal statistical rates, in: J.~Dy, A.~Krause (Eds.), Proceedings
  of the 35th International Conference on Machine Learning, Vol.~80 of
  Proceedings of Machine Learning Research, PMLR, 2018, pp. 5650--5659.

\bibitem{li2020learning}
S.~Li, Y.~Cheng, W.~Wang, Y.~Liu, T.~Chen, Learning to detect malicious clients
  for robust federated learning, arXiv preprint arXiv:2002.00211.

\bibitem{sun2019can}
Z.~Sun, P.~Kairouz, A.~T. Suresh, H.~B. McMahan, Can you really backdoor
  federated learning?, arXiv preprint arXiv:1911.07963.

\bibitem{naseri2020local}
M.~Naseri, J.~Hayes, E.~De~Cristofaro, Local and central differential privacy
  for robustness and privacy in federated learning, arXiv preprint
  arXiv:2009.03561.

\bibitem{zhao2018federated}
Y.~Zhao, M.~Li, L.~Lai, N.~Suda, D.~Civin, V.~Chandra, Federated learning with
  non-iid data, arXiv preprint arXiv:1806.00582.

\bibitem{xiong2021privacy}
Z.~Xiong, Z.~Cai, D.~Takabi, W.~Li, Privacy threat and defense for federated
  learning with non-i.i.d. data in aiot, IEEE Transactions on Industrial
  Informatics 18~(2) (2022) 1310--1321.
\newblock \href {http://dx.doi.org/10.1109/TII.2021.3073925}
  {\path{doi:10.1109/TII.2021.3073925}}.

\bibitem{byrd2012sample}
R.~H. Byrd, G.~M. Chin, J.~Nocedal, Y.~Wu, Sample size selection in
  optimization methods for machine learning, Mathematical programming 134~(1)
  (2012) 127--155.
\newblock \href {http://dx.doi.org/10.1007/s10107-012-0572-5}
  {\path{doi:10.1007/s10107-012-0572-5}}.

\bibitem{lecun2010mnist}
Y.~LeCun, C.~Cortes, C.~Burges, Mnist handwritten digit database, ATT Labs
  [Online]. Available: http://yann.lecun.com/exdb/mnist 2.

\bibitem{minka2000estimating}
T.~Minka, Estimating a dirichlet distribution (2000).

\bibitem{shayan2018biscotti}
M.~Shayan, C.~Fung, C.~J. Yoon, I.~Beschastnikh, Biscotti: A ledger for private
  and secure peer-to-peer machine learning, arXiv preprint arXiv:1811.09904.

\bibitem{abadi2016deep}
M.~Abadi, A.~Chu, I.~Goodfellow, H.~B. McMahan, I.~Mironov, K.~Talwar,
  L.~Zhang, Deep learning with differential privacy, in: Proceedings of the
  2016 ACM SIGSAC Conference on Computer and Communications Security,
  Association for Computing Machinery, New York, NY, USA, 2016, pp. 308--318.
\newblock \href {http://dx.doi.org/10.1145/2976749.2978318}
  {\path{doi:10.1145/2976749.2978318}}.

\end{thebibliography}

\newpage
\appendix

\section{Proof of Proposition \ref{proposition}}
\label{section::appendix}
\begin{proof}
\begin{footnotesize}
\begin{flalign}
&\|w_{k}^{T, t} - w_{cen}^{T, t}\| \nonumber \\
= &\big\|w^{T, t-1}_{k} - \eta\underbrace{\sum_{c=1}^{C} p_{k}(y=c) \nabla \mathbb{E}_{x \in D_k|y=c}[\log f_c(x; w^{T, t-1}_{k})]}_{A_1} \nonumber\\
& -w^{T, t-1}_{cen} + \eta\underbrace{\sum_{c=1}^{C} p(y=c) \nabla \mathbb{E}_{x \in D|y=c}[\log f_c(x; w^{T, t-1}_{cen})]}_{A_2}  \nonumber\\
& - \eta \underbrace{\sum_{c=1}^{C} p_k(y=c)\nabla \mathbb{E}_{x \in D|y=c}[\log(f_c(x; w_{cen}^{T, t-1})]}_{A_3}  + \eta  \underbrace{\sum_{c=1}^{C} p_k(y=c)\nabla \mathbb{E}_{x \in D|y=c}[\log(f_c(x; w_{cen}^{T, t-1})]}_{A_3}\big\| \\
 \overset{(1)}\leq & \|w^{T, t-1}_{k} - w^{T, t-1}_{cen}\| + \eta \|A_1-A_3\| + \eta \|A_2-A_3\| \\
= & \|w^{T, t-1}_{k} - w^{T, t-1}_{cen}\| + \eta \big \| \sum_{c=1}^{C} p_k(y=c) \big[\nabla \mathbb{E}_{x \in D_k|y=c}[\log f_c(x; w^{T, t-1}_{k})]-\nabla \mathbb{E}_{x \in D|y=c}[\log(f_c(x; w^{T, t-1}_{cen})]\big] \big\| \nonumber \\
& + \eta \big \| \sum_{c=1}^{C} \big[(p(y=c) -p_{k}(y=c)\big] \nabla \mathbb{E}_{x \in D|y=c}[\log(f_c(w^{T, t-1}_{k})] \big\|,
\end{flalign}
\end{footnotesize}
where the inequality $(1)$ holds due to the Cauchy–Schwarz inequality. By induction, we have:

\begin{footnotesize}
\begin{flalign}
&\|w_{k}^{T+1} - w_{cen}^{T+1}\| \nonumber\\
 \leq &\|w_{k}^{T} - w_{cen}^{T}\| + \eta \sum_{\tau=1}^t \big \| \sum_{c=1}^{C} \big[(p(y=c) -p_{k}(y=c)\big] \nabla \mathbb{E}_{x \in D|y=c}[\log(f_c(w^{T, \tau-1}_{k})] \big\| \nonumber\\
&  + \eta \sum_{\tau=1}^t  \big \| \sum_{c=1}^{C} p_k(y=c)\big[\nabla \mathbb{E}_{x \in D_k|y=c}[\log f_c(x; w^{T, \tau-1}_{k})]-\nabla \mathbb{E}_{x \in D|y=c}[\log(f_c(x; w^{T, \tau-1}_{cen})]\big] \big\|.
\end{flalign}
\end{footnotesize}

Hence, Eq. \ref{EQ_proposition_1} has been proved. And the proof of Eq. \ref{EQ_proposition_2} follows similar steps, and hence we omit the proof here. 
\end{proof} 

\end{document}